\def\CleanVersion{1}
\PassOptionsToPackage{table}{xcolor}
\documentclass[acmsmall,screen]{acmart}

\AtBeginDocument{
  }

\setcopyright{acmlicensed}
\copyrightyear{2026}
\acmYear{2026}
\acmJournal{TOIS}
\acmDOI{}
\setcopyright{none}
\makeatletter
\renewcommand{\@formatdoi}[1]{%
  \ifx\@acmDOI\@empty\else\url{https://doi.org/#1}\fi}
\makeatother
\makeatletter
\AddToHook{begindocument/end}{%
  \fancyfoot[RO,LE]{\footnotesize \@journalNameShort}%
  \fancypagestyle{firstpagestyle}{%
    \fancyhf{}%
    \renewcommand{\headrulewidth}{\z@}%
    \renewcommand{\footrulewidth}{\z@}%
    \fancyfoot[RO,LE]{\footnotesize \@journalNameShort}%
  }%
  \let\footnotetextcopyrightpermission\@gobble
}
\makeatother

\usepackage{needspace}
\usepackage{enumitem}
\usepackage{array}
\usepackage{float}
\usepackage[section]{placeins}

\usepackage{xcolor}
\newcommand{\rev}[1]{\textcolor{red}{#1}}
\ifdefined\CleanVersion
  \colorlet{red}{black}
\fi

\usepackage{multirow}
\usepackage{tabularx}
\usepackage{booktabs}

\definecolor{TableGroup}{HTML}{EFEFEF}
\definecolor{TableSeries}{HTML}{F7F8FA}
\definecolor{TableFocus}{HTML}{EDF2F6}
\definecolor{TableAux}{HTML}{6F6F6F}
\definecolor{TableTopOne}{HTML}{FEC6C6}
\definecolor{TableTopTwo}{HTML}{D0D0FE}
\definecolor{TableTopThree}{HTML}{C6FEC6}
\newcommand{\tablegrouprow}{\rowcolor{TableGroup}}

\newcommand{\tablefocusrow}{\rowcolor{TableFocus}}

\usepackage{bm}
\usepackage{amsmath}

\usepackage[ruled,vlined]{algorithm2e}

\usepackage{graphicx}
\usepackage{subcaption}
\newcommand{\revisiontable}{%
  \color{red}%
  \captionsetup{labelfont={color=red},textfont={color=red}}}

\usepackage{tcolorbox}
\usepackage[export]{adjustbox}
\tcbuselibrary{skins,breakable}

\AfterEndPreamble{
  \hypersetup{
    colorlinks=true,
    linkcolor=blue,
    citecolor=blue,
    urlcolor=blue,
    unicode=true,
    pdfborder={0 0 0}
  }
}

\usepackage{cleveref}

\newcommand{\punc}[1]{\hspace{0.7pt}\scalebox{0.82}{$\pm #1$}}
\newcommand{\presult}[2]{#1\punc{#2}}
\newcommand{\pbest}[2]{\textbf{#1}\punc{#2}}
\newcommand{\psecond}[2]{\underline{#1}\punc{#2}}
\newcommand{\psig}[1]{\textsuperscript{\textcolor{red}{\normalfont\scalebox{0.88}{#1}}}\hspace{0.07em}}

\Crefformat{section}{Section \textcolor{blue}{#2#1#3}}
\Crefformat{figure}{Figure \textcolor{blue}{#2#1#3}}
\Crefformat{subfigure}{Figure \textcolor{blue}{#2#1#3}}
\Crefformat{table}{Table \textcolor{blue}{#2#1#3}}
\Crefformat{equation}{Equation \textcolor{blue}{#2(#1)#3}}

\crefformat{cite}{\textcolor{blue}{#2#1#3}}
\Crefformat{cite}{Reference \textcolor{blue}{#2#1#3}}
\crefformat{cites}{\textcolor{blue}{#2#1#3}}
\Crefformat{cites}{References \textcolor{blue}{#2#1#3}}

\crefname{equation}{Eq.}{Eqs.}
\crefname{section}{Sec.}{Secs.}

\let\oldbibliography\thebibliography
\renewcommand{\thebibliography}[1]{%
  \oldbibliography{#1}%
  \setlength{\itemsep}{0pt plus 0.5pt}%
  \setlength{\parsep}{0pt}%
}

\begin{document}
\raggedbottom

\title[Modality-Guided Mixture of Structured Experts with Entropy-Triggered Routing for Multimodal Recommendation]{Modality-Guided Mixture of \rev{Structured Experts} with Entropy-Triggered Routing for Multimodal Recommendation}

\author{Ji Dai}
\email{daiji@bupt.edu.cn}
\affiliation{
  \institution{Beijing University of Posts and Telecommunications}
  \city{Beijing}
  \country{China}
}

\author{Quan Fang}
\authornote{Corresponding author: Quan Fang.}
\email{qfang@bupt.edu.cn}
\affiliation{
 \institution{Beijing University of Posts and Telecommunications}
 \city{Beijing}
     \country{China}}

\author{DeSheng Cai}
\email{caidsml@gmail.com}
\affiliation{%
  \institution{Tianjin University of Technology}
  \city{Tianjin}
  \country{China}}

\renewcommand{\shortauthors}{Ji Dai et al.}

\begin{abstract}
Multimodal recommenders combine collaborative behavior with visual and textual item evidence, but the relative usefulness of these sources can vary across individual user--item interactions. Independently trained source-specific diagnostic probes partition held-out interactions into behavior-, appearance-, semantics-, and mixed-evidence regimes across five benchmarks, within which capacity-matched fixed fusion rules exhibit systematic regime-dependent performance crossovers. This diagnostic observation motivates \textbf{MAGNET}, a multimodal graph recommender built around two core mechanisms supported by complementary evidence preparation. First, a calibrated expert bank organizes trainable experts by anchor source (behavior, appearance, or semantics) and fusion family (dominant, balanced, or complementary), while an interaction-conditioned router \rev{selects a sparse composition using all three evidence sources}. Second, an entropy-triggered, coverage-aware progressive schedule \rev{decouples population-level routing-mass coverage from per-instance decisiveness, transitioning from broad routing exploration to confident specialization once sufficient coverage is sustained, while preserving coverage thereafter}. As supporting evidence preparation, MAGNET separately encodes the observed interaction graph and a filtered content-induced structural view, applying cross-view alignment after independent propagation. We evaluate MAGNET on four Amazon domains and the \rev{non-Amazon MicroLens-100K benchmark, including tail-item and low-history warm-start user evaluation alongside matched expert-design controls}. Under the shared-feature, fixed-split protocol, MAGNET-DV exceeds the strongest protocol-compatible non-MAGNET baseline in every reported main-table cell, \rev{supported by seed-wise difference tests}. Both the fixed and KL-anchored structured variants achieve higher five-seed mean NDCG@20 than matched homogeneous and free-mixture alternatives across all five datasets. MAGNET additionally \rev{supports route-level diagnostics through explicit expert semantics}. Code and reproducibility configurations are available at \url{https://github.com/jidaivita/MAGNET}.
\end{abstract}

\begin{CCSXML}
<ccs2012>
   <concept>
       <concept_id>10002951.10003317.10003347.10003350</concept_id>
       <concept_desc>Information systems~Recommender systems</concept_desc>
       <concept_significance>500</concept_significance>
   </concept>
   <concept>
       <concept_id>10002951.10003227.10003251</concept_id>
       <concept_desc>Information systems~Multimedia information systems</concept_desc>
       <concept_significance>500</concept_significance>
   </concept>
   <concept>
       <concept_id>10010147.10010257.10010293.10010294</concept_id>
       <concept_desc>Computing methodologies~Neural networks</concept_desc>
       <concept_significance>300</concept_significance>
   </concept>
</ccs2012>
\end{CCSXML}

\ccsdesc[500]{Information systems~Recommender systems}
\ccsdesc[500]{Information systems~Multimedia information systems}
\ccsdesc[300]{Computing methodologies~Neural networks}

\keywords{Multimodal recommendation, graph neural networks, mixture of experts, adaptive routing, routing transparency}

 \maketitle

\section{Introduction}

\begin{figure}[!tp]
    \centering
    \includegraphics[width=\textwidth]{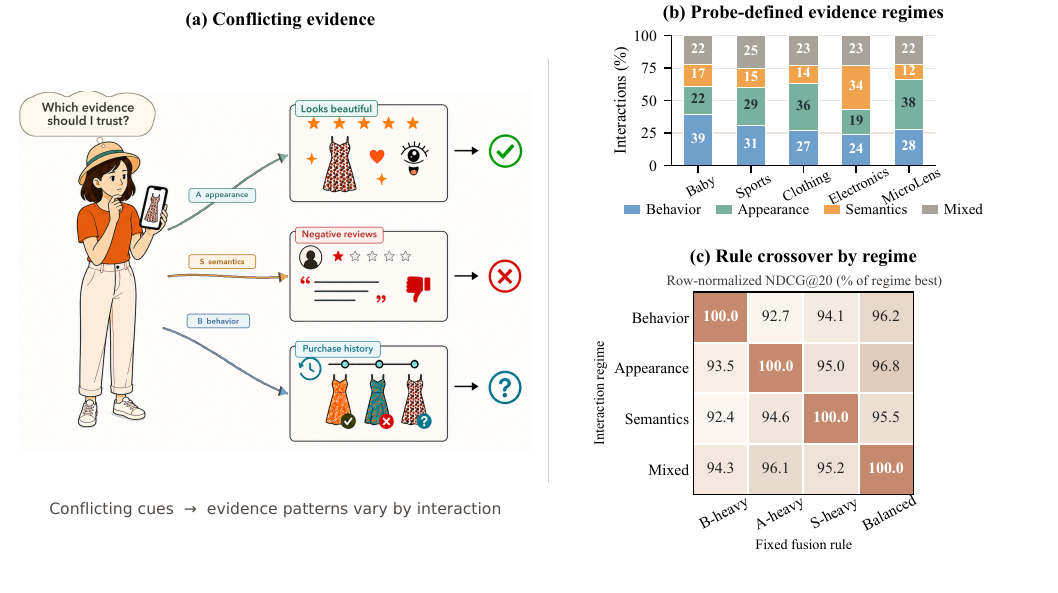}
    \Description{A conceptual shopping decision in which appearance, semantic feedback, and purchase history suggest different outcomes, followed by probe-defined evidence regime proportions across five datasets and a row-normalized NDCG at 20 matrix comparing four fixed fusion rules within four probe-defined regimes.}
    \caption{From conflicting cues to measurable fusion heterogeneity. \textbf{(a)} An illustrative user--item decision where appearance, semantic feedback, and purchase history suggest different preferences; the schematic is conceptual. \textbf{(b)} Held-out interaction proportions partitioned into four probe-defined evidence regimes across five datasets by independently trained source-specific diagnostic probes (behavior-, appearance-, and semantics-only). \textbf{(c)} Row-normalized NDCG@20 of four independently trained, capacity-matched fixed fusion rules evaluated within these probe-defined regimes. A value of 100 denotes the best fixed rule in that stratum; non-winning rules retain 92.4--96.8\% of regime-best performance, exhibiting systematic crossovers that provide diagnostic motivation for adaptive fusion. The diagnostic protocol and matched full-model controls are detailed in \Cref{sec:diagnostic-protocol,sec:template-design-analysis}, respectively.}
    \label{fig:motivation}
\end{figure}

Multimodal recommendation~\cite{Pan2026Multimodal} models user preference from behavioral interactions alongside item-side evidence such as images and text. These multimodal sources are especially valuable under sparse feedback, long-tail catalogs, and rapidly emerging items, but their relative usefulness is rarely uniform across all user--item decisions. A user may adhere to a stable behavioral habit in one interaction, rely predominantly on visual appeal or textual attributes in another, or balance multiple comparable cues. As illustrated conceptually in \Cref{fig:motivation}(a), visual styling, textual sentiment, and collaborative history can point toward conflicting recommendations.

This interaction-level variation can be empirically verified through controlled diagnostics. As shown in \Cref{fig:motivation}(b)--(c), partitioning held-out interactions by independently trained diagnostic probes reveals that capacity-matched fixed fusion rules suffer from systematic, regime-dependent performance crossovers. These diagnostic crossovers show that different interaction regimes favor different evidence mixtures, motivating interaction-conditioned fusion.

This interaction-level heterogeneity exposes a fundamental limitation of prevailing architectures. While multimodal graphs~\cite{Zhang00WWW21,ZhouS23,Ong2025Spectrum} and self-supervised refinement~\cite{Wu2021Self,Yu2022Graph,Hao2025ITCoHD} have significantly improved representation quality, they typically channel behavioral, visual, and semantic signals through a shared, static fusion pathway. Such coupling often leads to representational entanglement~\cite{Wang2020Disentangled,Liu2020Decoupled} and severe optimization imbalance across modalities~\cite{Peng2022Balanced,Wu2022Characterizing,Zhou2023Bootstrap}: an easily optimized or high-norm modality can dominate the fused representation even when an alternative source provides more decisive evidence for a specific decision.

Mixture-of-Experts (MoE) architectures and conditional computation~\cite{Shazeer2017Outrageously,Jiang2024Med,Ma2018Modeling} offer an intuitive foundation for interaction-conditioned fusion, yet standard homogeneous MoEs remain ill-suited for multimodal recommendation. Generic feed-forward experts lack explicit fusion semantics~\cite{Li2025UniMoE}, providing no structural guarantee that different activated branches represent diverse or meaningful modality trade-offs. Furthermore, sparse routing is prone to severe optimization instability: unconstrained routing easily collapses onto a subset of branches during early epochs, whereas static load-balancing objectives or a single global entropy statistic fail to reconcile early broad exploration with late-stage decisive specialization.

To resolve these challenges, we propose \textbf{MAGNET}, a multimodal graph recommender organized around two core mechanisms, supported by complementary structural evidence preparation. First, rather than relying on unconstrained branches, MAGNET establishes a calibrated bank of \rev{structured experts} organized across three evidence anchors (behavior, appearance, and semantics) and three canonical fusion families (dominant, balanced, and complementary); a lightweight router \rev{conditioned jointly on all three evidence sources} then dynamically selects a sparse composition per interaction. Second, to stabilize training dynamics, an entropy-triggered progressive schedule \rev{explicitly decouples population-level routing-mass coverage from per-instance routing decisiveness}, transitioning from broad routing-mass coverage to confident specialization only after coverage is reliably sustained. As supporting evidence preparation, MAGNET separately encodes the observed interaction graph and a filtered content-induced structural view, combining them via cross-view alignment while preserving the observed collaborative topology.

Our main contributions are summarized as follows:
\begin{itemize}[leftmargin=*,itemsep=2pt,topsep=2pt]
\item[--] \textbf{Structured fusion space with interaction-conditioned all-source routing.} We establish an interpretable expert bank spanning dominant, balanced, and complementary fusion families with calibrated inputs, accompanied by a router \rev{conditioned jointly on behavioral, visual, and semantic evidence}. \rev{Matched controls show that both structured variants achieve higher five-seed mean performance than homogeneous and unconstrained learnable alternatives across all five datasets.}
\item[--] \textbf{Coverage-aware progressive specialization schedule.} We formulate a \rev{dual-entropy optimization objective that separates population-level coverage from per-instance decisiveness}. The schedule promotes broad routing-mass coverage during early training and transitions via a persistent coverage trigger to confident specialization under an explicit coverage guard.
\item[--] \textbf{Rigorous cross-domain empirical validation.} Across four Amazon benchmarks and the \rev{non-commerce MicroLens-100K dataset}, MAGNET-DV exceeds the strongest protocol-compatible non-MAGNET baseline in every reported main-table cell, with \rev{cell-wise seed-difference tests supporting these comparisons}. Tail-item and low-history warm-start evaluations, together with route-level diagnostics, further characterize recommendation performance and transparent routing behavior.
\end{itemize}

\section{Related Work}

\subsection{From Collaborative Filtering to Graph-Based Multimodal Recommendation}
Multimodal recommendation systems (MMRS) have progressed from matrix factorization with side information to graph-based models that better cope with sparse feedback in Collaborative Filtering (CF)~\cite{Wang2019Neural}. Early factorization methods inject visual or textual signals into latent factors. VBPR~\cite{He2016VBPR} introduces CNN-derived visual features into the Bayesian Personalized Ranking (BPR) objective~\cite{Rendle2009BPR}. \rev{DeepCoNN~\cite{Zheng2017Joint} learns review-based user and item representations with parallel CNNs, whereas VECF~\cite{Chen2019Personalized} uses multimodal attention to model fine-grained visual preferences.} These approaches are effective under sparsity but do not explicitly capture higher-order user--item connectivity. Graph Neural Networks (GNNs) address this by propagating messages on the interaction graph; NGCF~\cite{Wang2019Neural} and LightGCN~\cite{He2020LightGCN} show strong gains from graph-based collaborative modeling. Multimodal GNNs such as MMGCN~\cite{Wei2019MMGCN} and GRCN~\cite{Wei2020Graph} further incorporate modality-specific propagation and content-guided graph refinement. Recent work shifts toward latent structure learning and denoising to mitigate noisy or missing edges: LATTICE~\cite{Zhang00WWW21} and SLMRec~\cite{Tao2023Self} mine implicit item--item relations, FREEDOM~\cite{ZhouS23} freezes reliable structures for efficiency, and SMORE~\cite{Ong2025Spectrum} performs frequency-domain fusion for noise handling in heterogeneous graphs. \rev{COHESION~\cite{Xu2025COHESION} couples composite graph convolution with early and late fusion so that representation learning and modality fusion reinforce one another.} \rev{VI-MMRec~\cite{Xu2026VIMMRec} takes a complementary plug-in perspective: it constructs similarity-aware virtual user--item interactions and can augment an existing recommender without modifying its core training architecture. MAGNET shares the broad intuition that content similarity can recover useful connectivity, but it filters candidate edges, preserves the observed and augmented structures as separate propagation views, and combines their representations after independent encoding. This separation is important when raw modality features are noisy because the observed graph remains independently encoded and the single-view variant remains a valid fallback.} Despite these advances, many methods still adopt coupled fusion pathways, which can entangle representations across modalities.

\subsection{Contrastive and Self-Supervised Learning in Recommendation}
\rev{In collaborative filtering more broadly}, SGL~\cite{Wu2021Self} and SimGCL~\cite{Yu2022Graph} construct auxiliary supervision by augmenting graph structure or perturbing embeddings, and train the model to keep user/item representations consistent across views. This idea extends naturally to multimodal recommendation by contrasting content modalities with collaborative signals. BM3~\cite{Zhou2023Bootstrap} streamlines training with a sufficiency-and-disentanglement objective that avoids negative sampling while aligning modality features. Subsequent work tailors contrastive objectives to multimodal challenges: MGCN~\cite{Yu2023MM} and MENTOR~\cite{Xu2025MENTOR} couple behavioral guidance with modality-specific views; GUME~\cite{Lin2024GUME} transfers modality information from head to tail items; SOIL~\cite{Su2024SOIL} captures second-order user interests; and ITCoHD-MRec~\cite{Hao2025ITCoHD} combines topological pruning with modal contrast to reduce noise propagation. A potential limitation of tightly aligned representations is that they can suppress complementary information or amplify noise when content cues conflict with user behavior.

\subsection{Structured Experts and Progressive Routing}
The Mixture of Experts (MoE) architecture was introduced to scale model capacity through conditional computation~\cite{Dai2024DeepSeekMoE,Fedus2022Switch}, and it is now widely used to cope with heterogeneous data and competing optimization signals. In recommender systems, MMoE~\cite{Ma2018Modeling} and PLE~\cite{Tang2020Progressive} employ gating to balance shared and task-specific experts, reducing negative transfer in multi-task learning (MTL). The same divide-and-conquer idea has since been applied to multimodal and graph settings. In vision--language models, V-MoE~\cite{Riquelme2021Scaling} and LIMoE~\cite{Mustafa2022Multimodal} show that sparse routing can scale contrastive learning across modalities. In graph learning, GraphMoRE~\cite{Guo2025GraphMoRE} allocates experts to different neighborhood ranges, \rev{Graph Mixture of Experts (GMoE)~\cite{Wang2023GraphMoE} routes individual nodes among aggregation experts with different hop ranges}, and MMOE~\cite{Yu2024MMoE} adapts MoE to multimodal interactions. Here MMOE denotes the multimodal-interaction model of Yu et al., distinct from the multi-task MMoE above. Overall, these studies suggest that specialization over distinct subspaces, whether topological or modal, improves representation expressiveness.

Adapting MoE to multimodal recommendation still faces two practical obstacles. First, experts are often generic feed-forward modules with limited semantic structure~\cite{Li2025UniMoE}, so a routed decision is hard to attribute to visual, textual, or collaborative evidence. Methods such as DMRL~\cite{Liu2023Disentangled} model modality-specific user interests, but do not explicitly organize an expert bank to cover interpretable fusion patterns. Second, sparse routing can collapse to a small subset of experts, producing severe load imbalance~\cite{Fedus2022Switch}. Standard load-balancing losses help, but their static form treats training uniformly even though early optimization requires broad utilization and later optimization benefits from sharper specialization. Entropy-aware optimization in recent reasoning models reports related observations about maintaining exploration while controlling excessive uncertainty~\cite{Dong2025Entropy,Cheng2025Reasoning}. \rev{These studies provide cross-domain evidence for the exploration--specialization distinction; MAGNET instantiates it for sparse expert routing by separating population-level routing-mass coverage from per-instance confidence.}

\section{Methods}
\label{sec:methods}

%========================
% Table: Summary of Key Notations (MAGNET)
%========================
\begin{table}[t]
\centering
\caption{Summary of key notations.}
\label{tab:notation}
\vspace{-2mm}
\scriptsize
\setlength{\tabcolsep}{7pt}
\renewcommand{\arraystretch}{0.96}
\resizebox{\linewidth}{!}{%
\begin{tabular*}{\linewidth}{@{\extracolsep{\fill}} l p{0.79\linewidth} @{}}
\toprule
\textbf{Notation} & \textbf{Description} \\
\midrule

\tablegrouprow
\multicolumn{2}{l}{\textbf{Sets and Graph Structures}} \\
$\mathcal{U},\mathcal{I}$ & Sets of users and items; sizes $|\mathcal{U}|$ and $|\mathcal{I}|$. \\
\(\mathcal{E},\mathcal{E}^{+}\) &
Observed interaction edges \(\mathcal{E}\), induced edges \(\mathcal{E}^{+}\) for the augmented view.\\
\(\mathcal{B}, b\) &
A mini-batch \(\mathcal{B}\subset\mathcal{E}\) and batch size \(b:=|\mathcal{B}|\).\\

$\mathcal{I}_u,$ $\mathcal{C}_{\mathrm{aug}}(u)$ & Training-only history of user $u$ and its augmentation candidate set. \\
\rev{$\mathcal{C}_{\mathrm{diag}}(u)$} & \rev{Candidate set used only by the independent motivating diagnostic.} \\
\(G_{UI}, G_{UIG}\) &
Original user--item graph \((\mathcal{U}\cup\mathcal{I},\mathcal{E})\) and augmented view \((\mathcal{U}\cup\mathcal{I},\mathcal{E}\cup\mathcal{E}^{+})\).\\

\(\mathbf{R}\) & User--item interaction matrix of \(G_{UI}\), \(\mathbf{R} \in \{0,1\}^{|\mathcal{U}|\times|\mathcal{I}|}\).\\

\(\mathcal{N}(i), k; r\) &
Modality-derived neighbor set \(\mathcal{N}(i)\) of item \(i\) (top-\(k\)) and the maximum retained augmentation candidates per user \(r\).\\

\(\mathcal{M}\) &
Content-modality set (default \(\mathcal{M}=\{\mathsf{A},\mathsf{S}\}\), i.e., visual/text).\\

\addlinespace[2pt]
\tablegrouprow
\multicolumn{2}{l}{\textbf{Features and Representations}} \\

$\mathbf{x}_i^{m},\mathbf{X}^{m}$ & Modality-$m$ feature of item $i$ and its matrix over all items. \\
$\mathbf{e}_u^{(\ell)},\mathbf{e}_i^{(\ell)}$ & User/item embeddings at graph layer $\ell$; aggregated to structural embeddings. \\

\(z_u^{UI}, z_i^{UI};\ z_u^{UIG}, z_i^{UIG}\) & View-specific behavior embeddings encoded on \(G_{UI}\) and \(G_{UIG}\).\\
\(z_u, z_i\) & Fused structural behavior embeddings from the graph view(s).\\

$\mathbf{h}_u^{m},\mathbf{h}_i^{m}$ & Modality cues for user/item (history pooling and/or feature projection), $m\in\mathcal{M}$. \\
\rev{$\mathbf r_u^B,\mathbf r_i^B$} & \rev{Projected behavioral cues $f_B(\mathbf z_u)$ and $f_B(\mathbf z_i)$.} \\
\rev{$s^{(e)}_{ui}$,$\hat{y}_{ui}$} & \rev{Expert-specific scalar compatibility score and routed preference score.} \\
\addlinespace[2pt]
\tablegrouprow
\multicolumn{2}{l}{\textbf{Model Architecture and Capacity}}\\
$d, L$ & Embedding dimension and the number of message-passing layers in the graph encoder. \\

$E,K$ & Total number of experts and the number of activated experts (Top-$K$). \\
\addlinespace[2pt]
\tablegrouprow
\multicolumn{2}{l}{\textbf{MoE Routing and Experts}} \\
\rev{$\mathbf{q}_{ui}$,$\boldsymbol{\pi}_{ui}$} & \rev{Interaction-conditioned query formed from separately normalized behavior, appearance, and semantic slots, and its dense routing distribution.} \\
\(\Gamma_{ui}, \tilde{\pi}_{ui}\) & Activated expert index set \(\Gamma_{ui}=\mathrm{TopK}(\pi_{ui},K)\) and renormalized weights.\\

$\Theta_{rt},\{\Theta_e\}_{e=1}^{E}$ & Router parameters and expert parameters for each expert $e$. \\
\addlinespace[2pt]
\tablegrouprow
\multicolumn{2}{l}{\textbf{Entropy-Guided Schedule and Optimization}} \\
$\Theta$ & All trainable parameters of MAGNET (encoder/projections/router/experts). \\
$\alpha$, $\beta$, $\delta$ & Triplet-template controls for dominant/balanced/complementary expert compositions. \\
\rev{$\epsilon_{\mathrm{cal}}$} & \rev{Fixed numerical stabilizer ($10^{-8}$) for source-wise scale calibration.} \\
\rev{$\xi$} & \rev{Fixed minimal anchor mass ($0.10$) in complementary expert templates.} \\
\(\bar{\pi}, H(\cdot)\) & Batch-mean \emph{dense} routing and its entropy, \(\bar{\pi} = \frac{1}{|\mathcal{B}|} \sum_{(u,i)\in\mathcal{B}} \pi_{ui}\).\\

\rev{$H_{\mathrm{cov}},H_{\mathrm{inst}}$} & \rev{Normalized population coverage entropy $H(\bar{\pi})/\log E$ and mean normalized per-instance routing entropy.} \\

\rev{$I_{\mathrm{route}}$} & \rev{Routing mutual-information proxy, $H_{\mathrm{cov}}-H_{\mathrm{inst}}$.} \\

$N_{\mathrm{eff}}$ & Effective number of experts, $N_{\mathrm{eff}}=\exp\!\big(H(\bar{\boldsymbol{\pi}})\big)$. \\

$H^*, W$ & Coverage-entropy threshold and persistence window for stage switching. \\

\textit{stage}, n & Stage indicator and counter for consecutive steps with $H_{\mathrm{cov}}\ge H^*$. \\

$\mathcal{L}_{\mathrm{BPR}},\mathcal{L}_{ctr}$ & Pairwise BPR loss and view-contrastive loss (InfoNCE) on dual views (optional). \\

\(\rho\) & Negative sampling ratio in BPR training.\\

$p, T_{\max}, N$ & Early-stopping patience, maximum training epochs, and evaluation cutoffs (top-$N$). \\

\rev{\(L_{\mathrm{route}}^{(1)},L_{\mathrm{route}}^{(2)}\)} & \rev{Stage-specific routing objectives for coverage promotion and specialization with coverage preservation.}\\

\rev{$\lambda_r$} & \rev{Shared routing strength, with $(\lambda_{\rm exp},\lambda_{\rm cov},\lambda_{\rm MI},\lambda_{\rm guard})=\lambda_r(1,6,6,2)$.} \\

\rev{$T_{\rm exp},\lambda_{\rm anchor}$} & \rev{Transient Stage~1 exploration window $\lfloor0.05T\rfloor$ (in optimization updates) and KL-anchor weight for the anchored learnable control.} \\

$\eta, p_{\mathrm{d}}, \lambda, \lambda_{\mathrm{ctr}}, \tau$ & Learning rate, dropout rate, weight decay, contrastive weight, and temperature $\tau$. \\

\addlinespace[2pt]
\tablegrouprow
\multicolumn{2}{l}{\textbf{Variants and Abbreviations}} \\
$\mathrm{B},\mathrm{A},\mathrm{S}$ & Branch tags for Behavior/Appearance/Semantics expert groups. \\
$\mathrm{Dom},\mathrm{Bal},\mathrm{Com}$ & Template tags: dominant, balanced, and complementary expert compositions. \\

MAGNET-DV & Dual-view (default): use $\mathcal{G}_{UI}$+$\mathcal{G}_{UIG}$ for routing/scoring (optional $\mathcal{L}_{\mathrm{ctr}}$). \\
MAGNET-SV & Single-view: use only $\mathcal{G}_{UI}$ with $\mathcal{E}^{+}=\varnothing$ (same MoE and recommendation/routing objectives; no $\mathcal{L}_{\mathrm{ctr}}$). \\

\bottomrule
\end{tabular*}
}
\vspace{-3mm}
\end{table}

\subsection{Problem Setup and Overview}
\label{sec:problem}

\subsubsection{Task Definition and Notation}
\label{sec:task-notation}
We study implicit-feedback multimodal recommendation, where observed user--item interactions
provide the primary supervision signal, and item-side multimodal content offers complementary
evidence, especially under sparsity and long-tail regimes. Each item is associated with appearance
(visual) and semantic (text) features, while user-side modality profiles are typically unavailable.

Let $\mathcal{U}$ and $\mathcal{I}$ denote the sets of users and items, respectively, and let
$\mathcal{D}_{tr} \subseteq \mathcal{U}\times \mathcal{I}$ be the observed training interactions.
We denote the observed edge set as $\mathcal{E}=\{(u,i):(u,i)\in\mathcal{D}_{tr}\}$, and represent
implicit feedback by the binary interaction matrix $\mathbf{R}\in\{0,1\}^{|\mathcal{U}|\times|\mathcal{I}|}$,
where $\mathbf{R}_{ui}=1$ iff $(u,i)\in\mathcal{E}$.
For each user $u$, let $\mathcal{I}_u=\{i\in\mathcal{I}:(u,i)\in\mathcal{E}\}$ be the training-only
history.

Throughout the paper, we use \textbf{Behavior (B)} to denote interaction-induced collaborative signals,
and \textbf{Appearance (A)} and \textbf{Semantics (S)} to denote visual and textual modalities, respectively.
We write the content-modality set as $\mathcal{M}=\{A,S\}$ by default. Here, $\mathcal{M}$ designates item-side content modalities (visual and textual), while $B$ represents the behavioral collaborative source from observed interactions; the triplet $\{B,A,S\}$ thus treats collaborative behavior as a third evidence source alongside content modalities $\mathcal{M}$.

Each item $i$ is associated with raw modality features $\mathbf{x}_i^{m}$ for $m\in\mathcal{M}$.
The goal is to learn a scoring function $\hat{y}_{ui}$ for ranking items for each user.
Unless otherwise specified, all learned embeddings and hidden representations are $d$-dimensional.
\Cref{tab:notation} summarizes the key notation used throughout the method.

\begin{figure}[t]
    \centering
    \includegraphics[width=\textwidth]{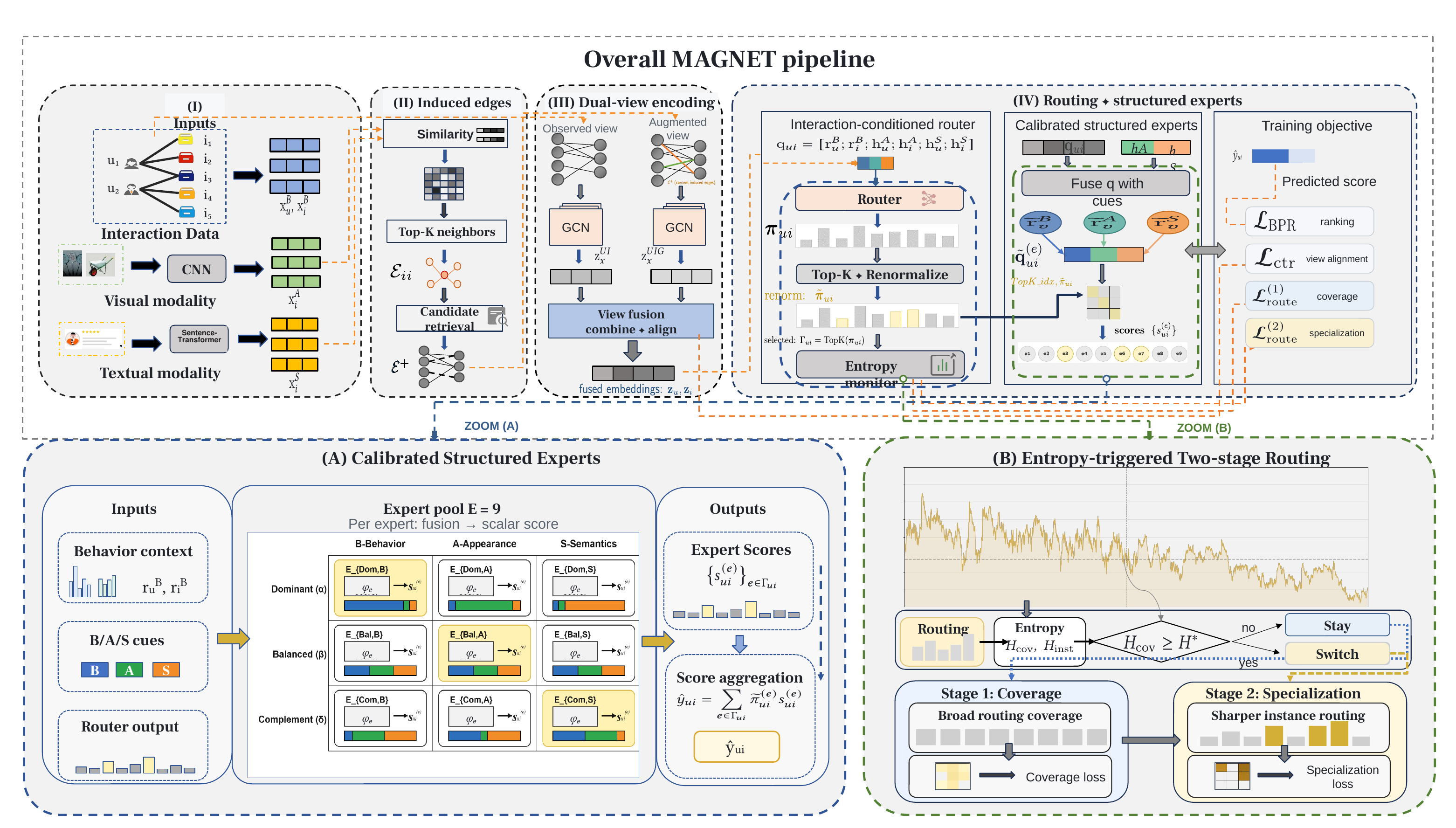}
    \Description{Overview of MAGNET, showing evidence preparation, dual-view graph encoding, calibrated structured experts, interaction-conditioned sparse routing, and the entropy-triggered two-stage training schedule.}
    \caption{\textbf{Overview of MAGNET.} The first row follows the end-to-end path from multimodal inputs and dual-view graph encoding to \rev{interaction-conditioned sparse routing over calibrated structured experts}. The second row exposes the two design principles: explicit coverage of complementary fusion patterns and \rev{an entropy-triggered transition from broad routing-mass coverage to per-instance specialization}.}
    \label{fig:main}
\end{figure}

\subsubsection{Model Overview}
\label{sec:overview}
As shown in \Cref{fig:main}, MAGNET consists of complementary structural evidence preparation supporting two core mechanisms:

\paragraph{Supporting evidence preparation (\Cref{sec:graph_construct}).}
Beyond the observed interaction graph, MAGNET constructs an augmented structural view from a filtered set of content-induced user--item candidate edges. Parallel dual-view encoding preserves historical collaborative topology while introducing content-informed connectivity for sparse and long-tail settings.

\begin{enumerate}[label=\textbf{\arabic*.}, leftmargin=1.5em]

\item \textbf{Interaction-conditioned structured fusion (\Cref{sec:magnet-moe}).}
MAGNET uses a sparse Mixture-of-Experts prediction head whose experts instantiate an interpretable triplet-template geometry over \{behavior, visual, text\} while retaining expert-specific trainable transformations. This defines explicit fusion patterns, with \rev{interaction-conditioned routing selecting experts from joint behavioral and modality evidence while each selected branch transforms and scores its corresponding composition}.

\item \textbf{Stable routing specialization (\Cref{sec:entropy}).}
MAGNET monitors routing entropy to govern a two-stage training schedule: \rev{an early coverage stage that promotes broad population-level routing mass, and a later specialization stage that sharpens routing once this mass is sufficiently dispersed}. This entropy-driven trigger \rev{coordinates} the transition from broad exploration to confident expert specialization.

\end{enumerate}

\subsection{Content-Induced Structural Augmentation and Dual-View Backbone}
\label{sec:graph_construct}
\subsubsection{Content-Guided Candidate Expansion}
\label{high-order-sets}
MAGNET leverages multimodal content to construct optional high-order structural signals, while keeping subsequent user--item message passing shallow.
Concretely, for each modality $m\in\{A,S\}$, we compute cosine item--item similarity
$s^{m}_{ij}=\langle \mathbf{x}^{m}_{i},\mathbf{x}^{m}_{j}\rangle/(\|\mathbf{x}^{m}_{i}\|\,\|\mathbf{x}^{m}_{j}\|)$ from raw item features.
For each item $i$, we retain only the top-$k$ most similar items under each modality to form sparse neighbor sets
$\mathcal{N}^{A}(i)$ and $\mathcal{N}^{S}(i)$.
For convenience, we write $\mathcal{N}(i)=\mathcal{N}^{A}(i)\cup \mathcal{N}^{S}(i)$ when the modality index is clear.

Given a user $u$ with training-only history $\mathcal{I}_u$, we expand a candidate pool by traversing item--item neighborhoods from the user history.
For each candidate item $j\notin \mathcal{I}_u$, we define a content-induced relevance score:
\begin{equation}
c_{u,j}=\sum_{i\in \mathcal{I}_u}\frac{s^{A}_{ij}+s^{S}_{ij}}{2},
\label{eq:augmentation-score}
\end{equation}
and retain at most the top-$r$ augmentation candidates $\mathcal{C}_{\mathrm{aug}}(u)$ according to $c_{u,j}$.

We denote the observed training edges by
$\mathcal{E}=\{(u,i):(u,i)\in \mathcal{D}_{tr}\}$ and the induced edges by
$\mathcal{E}^{+}=\{(u,j):u\in \mathcal{U},\, j\in \mathcal{C}_{\mathrm{aug}}(u)\}$.
Thus, $r$ serves as a per-user retention budget rather than a guarantee of $r$ new
edges: the realized edge set is determined after candidate filtering and edge
deduplication.

This step acts as \emph{graph augmentation} rather than multi-hop message passing on the user--item graph: it injects additional content-informed edges as optional structural evidence, which is intended to provide additional connectivity under sparse and long-tail settings.

\subsubsection{Dual-View Structural Graphs}
Based on the observed edge set $\mathcal{E}$ and the induced edges $\mathcal{E}^{+}$,
we construct two structural views:
the original user--item graph $\mathcal{G}_{UI}=(\mathcal{U}\cup \mathcal{I},\, \mathcal{E})$
and the augmented graph $\mathcal{G}_{UIG}=(\mathcal{U}\cup \mathcal{I},\, \mathcal{E}\cup \mathcal{E}^{+})$.
Dual-view (DV) is our default formulation: we encode behavioral
signals independently on $\mathcal{G}_{UI}$ and $\mathcal{G}_{UIG}$ to obtain view-specific behavior representations $\mathbf{z}^{UI}$ and $\mathbf{z}^{UIG}$, which are
then fused into $\mathbf{z}$ for downstream routing and scoring via fixed mean fusion, i.e.,
$\mathbf{z} \leftarrow \frac{1}{2}(\mathbf{z}^{UI}+\mathbf{z}^{UIG})$ (see \Cref{sec:Structural_Encoding}).
The router selects structured experts after this fixed mean fusion and does not gate or reweight the two structural views, avoiding additional fusion hyperparameters.
Since a simple mean assumes the two views occupy a comparable representation space, we optionally add a view-contrastive alignment loss $\mathcal{L}_{\mathrm{ctr}}$ (described in \Cref{sec:objective_routing_reg}) to regularize the representation gap between $\mathbf{z}^{UI}$ and $\mathbf{z}^{UIG}$, stabilizing the fused representation $\mathbf{z}$ for routing and scoring.

\textbf{Single-view (SV)} is a fallback variant that removes induced edges by setting $\mathcal{E}^{+}=\varnothing$,
so that $\mathcal{G}_{UIG}$ collapses to $\mathcal{G}_{UI}$ and only one structural view is encoded.
Unless stated otherwise, we report DV as the default model.

\subsubsection{Structural Encoding with Shallow Propagation}
\label{sec:Structural_Encoding}
MAGNET adopts LightGCN-style message passing to extract behavioral representations from each structural view.
We share the same trainable ID embeddings $\mathbf{e}^{(0)}_{u}$ and $\mathbf{e}^{(0)}_{i}$ across views,
and run shallow propagation on $\mathcal{G}_{UI}$ and $\mathcal{G}_{UIG}$ to obtain $\mathbf{z}^{UI}$ and $\mathbf{z}^{UIG}$, respectively.
The two variants share all downstream trainable modules; the augmented
adjacency in MAGNET-DV is stored as a non-trainable sparse graph and alters
propagation computation rather than trainable model capacity.

Let $\mathbf{e}^{(0)}_{u}$ and $\mathbf{e}^{(0)}_{i}$ be trainable ID embeddings for user $u$ and item $i$.
For each view $v\in\{UI, UIG\}$ and each layer $\ell=0,\dots,L-1$, we update user and item representations via:
\begin{equation}
\mathbf{e}^{(\ell+1),v}_{u}=\sum_{i\in N^{v}(u)}\frac{1}{\sqrt{|N^{v}(u)|\,|N^{v}(i)|}}\,\mathbf{e}^{(\ell),v}_{i},\quad
\mathbf{e}^{(\ell+1),v}_{i}=\sum_{u\in N^{v}(i)}\frac{1}{\sqrt{|N^{v}(i)|\,|N^{v}(u)|}}\,\mathbf{e}^{(\ell),v}_{u},
\end{equation}
where $N^{v}(u)$ and $N^{v}(i)$ denote the neighbor sets in graph $\mathcal{G}_{v}$. We use full-neighborhood propagation in both structural views: each layer multiplies the complete normalized sparse adjacency by the current node representations. No fanout-based neighbor sampling is used in the reported experiments.

We initialize $\mathbf{e}^{(0),v}_{u}\!=\!\mathbf{e}^{(0)}_{u}$ and $\mathbf{e}^{(0),v}_{i}\!=\!\mathbf{e}^{(0)}_{i}$ for both views.

Following LightGCN, we average the layer-wise representations within each view and then aggregate the two views:
\begin{equation}
\mathbf{z}^{v}_{x}=\frac{1}{L+1}\sum_{\ell=0}^{L}\mathbf{e}^{(\ell),v}_{x},\qquad
\mathbf{z}_{x}=\frac{1}{2}\left(\mathbf{z}^{UI}_{x}+\mathbf{z}^{UIG}_{x}\right),\qquad
x\in\{u,i\},\;v\in\{UI,UIG\}.
\end{equation}

\begin{algorithm}[t]
\caption{MAGNET Scoring via Modality-Guided MoE (Sparse Expert Activation)}
\label{alg:magnet-moe}
\footnotesize
\SetAlgoLined
\DontPrintSemicolon
\SetKwComment{tcp}{\ttfamily\scriptsize // }{}
\newcommand{\atcp}[1]{\tcp*[r]{#1}}

\SetKwInOut{Input}{Require}
\SetKwInOut{Output}{Ensure}

\Input{
Behavior embeddings $\mathbf{z}_u,\mathbf{z}_i$; modality cues $\{\mathbf{h}_u^m,\mathbf{h}_i^m\}_{m\in\mathcal{M}}$;\\
Router parameters $\Theta_{rt}$; expert parameters $\{\Theta_e\}_{e=1}^{E}$; activated experts $K$.
}
\Output{
Preference score $\hat{y}_{ui}$, dense routing $\boldsymbol{\pi}_{ui}$, activated expert set $\Gamma_{ui}$,
and renormalized weights $\tilde{\boldsymbol{\pi}}_{ui}$.
}

{\bf Step 0: Prepare modality cues}\;
\If{$\{\mathbf{h}_u^m\},\{\mathbf{h}_i^m\}$ are not precomputed}{
Compute modality cues $\{\mathbf{h}_u^m\}_{m\in\mathcal{M}}$ and $\{\mathbf{h}_i^m\}_{m\in\mathcal{M}}$\;
}
$\mathbf{r}_u^B\leftarrow f_B(\mathbf{z}_u)$\;
$\mathbf{r}_i^B\leftarrow f_B(\mathbf{z}_i)$\;

{\bf Step 1: Dense routing distribution}\;
\rev{$q_{ui}\leftarrow [\operatorname{Norm}(\mathbf{r}_u^B);\operatorname{Norm}(\mathbf{r}_i^B);$}
\rev{$\hspace{3.2em}\operatorname{Norm}(\mathbf{h}_u^A);\operatorname{Norm}(\mathbf{h}_i^A);\operatorname{Norm}(\mathbf{h}_u^S);\operatorname{Norm}(\mathbf{h}_i^S)]$}\;
$\boldsymbol{\pi}_{ui}\leftarrow \mathrm{Softmax}\!\left(\mathrm{Router}(q_{ui};\Theta_{rt})\right)$\;

{\bf Step 2: Sparse expert activation \& routed score aggregation}\;
$\Gamma_{ui}\leftarrow \mathrm{TopK}(\boldsymbol{\pi}_{ui},K)$\;
$\tilde{\boldsymbol{\pi}}_{ui}\leftarrow \mathrm{Renorm}(\boldsymbol{\pi}_{ui},\Gamma_{ui})$\;
$\hat{y}_{ui}\leftarrow 0$\;
\ForEach{$e\in\Gamma_{ui}$}{
$s^{(e)}_{ui}\leftarrow \mathrm{Expert}_e\!\left(\mathbf{r}_u^B,\mathbf{r}_i^B,\{\mathbf{h}_u^m\}_{m\in\mathcal{M}},\{\mathbf{h}_i^m\}_{m\in\mathcal{M}};\Theta_e\right)$\;
$\hat{y}_{ui}\leftarrow \hat{y}_{ui}+\tilde{\pi}^{(e)}_{ui}\cdot s^{(e)}_{ui}$\;
}
\Return $(\hat{y}_{ui},\boldsymbol{\pi}_{ui},\Gamma_{ui},\tilde{\boldsymbol{\pi}}_{ui})$\;
\vspace{2pt}
\end{algorithm}

In the single-view variant, $\mathcal{E}^{+}=\varnothing$ and we set $\mathbf{z}_{u}=\mathbf{z}^{UI}_{u}$ and $\mathbf{z}_{i}=\mathbf{z}^{UI}_{i}$.
We refer to the default dual-view setting as \textbf{MAGNET-DV} and the single-view ablation ($\mathcal{E}^{+}=\varnothing$) as \textbf{MAGNET-SV}.

\subsection{Triplet-Template Mixture-of-Experts for Multimodal Fusion}
\label{sec:magnet-moe}

MAGNET treats interaction-induced behavior as the primary supervision signal and uses item-side multimodal content as complementary evidence.
From the dual-view structural backbone (\Cref{sec:graph_construct}), we obtain fused behavior embeddings $\mathbf{z}_u, \mathbf{z}_i \in \mathbb{R}^{d}$ for each interaction $(u,i)$.
We then build a structured Mixture-of-Experts (MoE) layer. Each branch receives an interpretable triplet composition over $\{B,A,S\}$ and applies a trainable expert-specific transformation, while a router learns interaction-wise sparse gating over the expert bank. The templates organize qualitatively different fusion patterns; interaction-level adaptivity comes from routing, with an anchored learnable variant allowing bounded movement around the canonical templates. The entropy-triggered schedule in \Cref{sec:entropy} controls when routing transitions from broad routing-mass coverage to specialization.

For clarity, we summarize the full MoE-based scoring procedure (from routing to aggregation and prediction) in \Cref{alg:magnet-moe}, which will be reused in the training pipeline.

\subsubsection{Modality Cues: Item Projection and History-Induced User Cues}
Let $\mathcal{M}$ denote the item content modality set (default $\mathcal{M}=\{A,S\}$).
For each item $i$, a lightweight projection $f_m(\cdot)$ produces its modality cue. The supplied fixed visual and textual feature tensors follow the shared dataset preprocessing and enter these affine projections without an additional model-side $\ell_2$ normalization. Since user-side raw modality profiles are unavailable in implicit feedback, we average projected cues over the training-only history $\mathcal{I}_u$:
\begin{equation}
\label{eq:7}
\mathbf{h}_i^{m}= f_{m}(\mathbf{x}_i^{m}), \qquad
\mathbf{h}_u^{m}=\frac{1}{|\mathcal{I}_u|}\sum_{j\in \mathcal{I}_u} \mathbf{h}_j^{m},\qquad m\in\{A,S\}.
\end{equation}
This design keeps behavior embeddings $(\mathbf{z}_u, \mathbf{z}_i)$ as the only direct user signal, while allowing content evidence to participate through history-induced cues without leaking test-time information.
\rev{All user modality cues are candidate-independent averages over the fixed training-only history $\mathcal{I}_u$ and are not recomputed in leave-one-out form for individual positive pairs. The same cue is reused for both items in each BPR tuple. Thus, a positive contributes $1/|\mathcal{I}_u|$ only as an observed training interaction; validation and test targets never enter cue construction.}

\subsubsection{Triplet Experts with Shared Templates}
\label{sec:magnet-triplet-experts}
We abbreviate Behavior/Appearance/Semantics as B/A/S (\Cref{sec:task-notation})
and use the global order $[B,A,S]$. A shared affine map $f_B$ places the fused
structural embedding in the same $d$-dimensional evidence space as the content
cues:
\begin{equation}
\label{eq:magnet-ras}
\mathbf{r}_u^{B}=f_B(\mathbf{z}_u),\quad \mathbf{r}_i^{B}=f_B(\mathbf{z}_i),\qquad
\mathbf{r}_u^{A}=\mathbf{h}_u^{A},\ \mathbf{r}_i^{A}=\mathbf{h}_i^{A},\qquad
\mathbf{r}_u^{S}=\mathbf{h}_u^{S},\ \mathbf{r}_i^{S}=\mathbf{h}_i^{S}.
\end{equation}

\paragraph{Scale calibration.}
\rev{Behavioral, appearance, and semantic representations are produced by heterogeneous encoders and therefore need not have comparable norms. Directly mixing them would allow numerical scale to override a nominal template preference. Source calibration is applied after the behavioral/content projections and, for user-side content cues, after training-history pooling, but before template-based source fusion:
\begin{equation}
\widetilde{\mathbf r}^{m}_{v}=\operatorname{Cal}_{m}(\mathbf r^{m}_{v})
=\frac{\mathbf r^{m}_{v}}{\lVert\mathbf r^{m}_{v}\rVert_2+\epsilon_{\mathrm{cal}}},
\qquad m\in\{B,A,S\},\quad v\in\{u,i\},
\label{eq:cue-calibration}
\end{equation}
where $\epsilon_{\mathrm{cal}}$ is a numerical stabilizer. Calibration places the three inputs on a common norm scale so that the triplet coefficients govern their effective mixture.
We use $\epsilon_{\mathrm{cal}}=10^{-8}$ across all datasets and variants.}

Each expert is specified by a nonnegative source triplet $\mathbf{w}=[w^{B},w^{A},w^{S}]^\top\in\mathbb{R}_{\ge 0}^{3}$ with a normalized budget $\sum_{m\in\{B,A,S\}}w^{m}=1$. Given an expert triplet $\mathbf{w}$, we form expert-specific user and item representations by calibrated triplet fusion:
\begin{equation}
\label{eq:magnet-triplet-fusion}
\mathbf{x}_u(\mathbf{w})=\sum\nolimits_{m\in\{B,A,S\}} w^{m}\widetilde{\mathbf{r}}_u^{m},\qquad
\mathbf{x}_i(\mathbf{w})=\sum\nolimits_{m\in\{B,A,S\}} w^{m}\widetilde{\mathbf{r}}_i^{m}.
\end{equation}
An expert module then produces a scalar pairwise compatibility score through a
lightweight expert-specific head:
\begin{equation}
s^{(e)}_{ui}=\mathrm{Expert}_e\!\left(
\mathbf{r}^B_u,\mathbf{r}^B_i,\{\mathbf{h}^m_u\}_{m\in\mathcal{M}},\{\mathbf{h}^m_i\}_{m\in\mathcal{M}};\Theta_e
\right)
=\phi\!\left([\mathbf{x}_u(\mathbf{w}^{(e)});\mathbf{x}_i(\mathbf{w}^{(e)})];\Theta_e\right),
\label{eq:expert-pair}
\end{equation}
where $\mathbf{w}^{(e)}$ is the triplet of expert $e$ (defined below), and
$\phi(\cdot;\Theta_e)$ is a lightweight scalar-valued transformation. The
calibrated coefficients shape both forward composition and the corresponding
backward signal, while the expert-specific transformation parameters remain
trainable.

\paragraph{Core definitions: expert groups, expert types, and template instantiation.}
To make the expert pool explicit and interpretable, we formalize expert groups (by anchor source) and shared type-level templates.

\textbf{(1) Expert groups (anchor source).}
We organize experts into three evidence-source groups $g\in\{B,A,S\}$, where each group specifies an anchor source (following \Cref{sec:task-notation}).
The anchor is a semantic label for structuring and analyzing experts; it does \emph{not} hard-assign instances.
This design enables group-wise analysis (e.g., utilization by anchor) and yields a compact $3\times 3$ expert pool without introducing additional group-specific hyperparameters.

\textbf{(2) Expert types (how the anchor participates).}
Within each group, we instantiate three expert types $t\in\{\text{Dom},\text{Bal},\text{Com}\}$:
(i) \emph{Dominant (Dom)}: the anchor receives a large portion of the fusion budget, with auxiliaries down-weighted (and possibly approaching zero at the boundary);
(ii) \emph{Balanced (Bal)}: a smoother trade-off with a bounded anchor bias;
(iii) \emph{Complementary (Com)}: auxiliaries provide the major evidence while retaining a small \emph{anchor mass}.

\textbf{(3) Type-level templates in canonical order.}
A type-level template maps a scalar to a canonical triplet $\tilde{\mathbf{w}}^{t}(\cdot)\in\mathbb{R}^{3}$ under the order $[\text{anchor},\text{aux}_{1},\text{aux}_{2}]$:
\begin{equation}
\begin{aligned}
\tilde{\mathbf{w}}^{\text{Dom}}(\alpha)&=(1-\alpha)\!\left[\tfrac{1}{2},\tfrac{1}{4},\tfrac{1}{4}\right]+\alpha[1,0,0],\\
\tilde{\mathbf{w}}^{\text{Bal}}(\beta)&=\left[\tfrac{1}{3}+\tfrac{\beta}{6},\tfrac{1}{3}-\tfrac{\beta}{12},\tfrac{1}{3}-\tfrac{\beta}{12}\right],\\
\tilde{\mathbf{w}}^{\text{Com}}(\delta)&=\left[\xi,(1-\xi)\delta,(1-\xi)(1-\delta)\right],
\end{aligned}
\qquad \alpha,\beta,\delta\in[0,1],
\end{equation}
where $\beta$ controls the anchor bias of the balanced family, $\delta$ allocates mass between the two auxiliary sources of the complementary family, and $\xi>0$ is a small fixed constant encoding a minimal \emph{anchor mass}. We set $\xi=0.10$ across all datasets. When $g=B$, this corresponds to retaining a minimal behavior anchor, reflecting the inductive bias of implicit feedback;
when $g\in\{A,S\}$, it prevents the anchor content source from being entirely discarded.
We treat $\alpha,\beta,\delta$ as \emph{template hyperparameters}: they are type-level shared and fixed during training, and we analyze their sensitivity in \Cref{sec:hparam}.

\textbf{(4) From templates to a $3\times 3$ expert pool (group-specific instantiation).}
To obtain the final expert triplet in the global order $[B,A,S]$, we apply a fixed permutation operator $P_g(\cdot)$ that maps
the canonical order $[\text{anchor},\text{aux}_{1},\text{aux}_{2}]$ to $[B,A,S]$ according to group $g$:
\begin{equation}
\mathbf{w}^{g,t} = P_{g}\big(\tilde{\mathbf{w}}^{t}\big), \qquad g\in\{B,A,S\},\; t\in\{\text{Dom},\text{Bal},\text{Com}\}.
\end{equation}
We fix the auxiliary order as
$\operatorname{aux}(B)=(A,S)$,
$\operatorname{aux}(A)=(B,S)$, and
$\operatorname{aux}(S)=(B,A)$. Hence, for a canonical triplet
$(a,b,c)=[\mathrm{anchor},\mathrm{aux}_1,\mathrm{aux}_2]$,
\begin{equation}
P_B(a,b,c)=(a,b,c),\qquad
P_A(a,b,c)=(b,a,c),\qquad
P_S(a,b,c)=(b,c,a),
\label{eq:expert-permutations}
\end{equation}
where every output follows the global $[B,A,S]$ order.
$P_g(\cdot)$ is deterministic. We index experts by $e\leftrightarrow(g,t)$, yielding $E=9$ experts by default. The resulting canonical bank spans three interpretable fusion patterns: Dom represents single-source dominance, Bal represents coordinated use of multiple sources, and Com represents auxiliary correction under a retained anchor. \Cref{sec:template-design-analysis} compares this compact structured basis with matched homogeneous and learnable alternatives.

\paragraph{Fixed and anchored learnable templates.}
The default fixed variant uses $\mathbf w^{(e)}=\mathbf w^{g,t}$ throughout training, preserving the semantic identity of every branch. To test whether exact fixation is necessary, we introduce an \rev{anchored learnable variant} with logits $\bm\rho^{(e)}$ initialized from the canonical template:
\begin{equation}
\mathbf w_{\mathrm{learn}}^{(e)}=\operatorname{Softmax}(\bm\rho^{(e)}),\qquad
\mathcal L_{\mathrm{anchor}}=\frac{1}{E}\sum_{e=1}^{E}\operatorname{KL}\!\left(\mathbf w_{\mathrm{learn}}^{(e)}\,\middle\|\,\mathbf w_{0}^{(e)}\right).
\label{eq:anchored-template}
\end{equation}
This variant permits data-driven adjustment while retaining a transparent reference geometry. \rev{It uses the forward KL direction shown above with $\lambda_{\mathrm{anchor}}=0.05$. Free learnable mixtures use the same learnable-softmax triplet parameterization with $\lambda_{\mathrm{anchor}}=0$.} Fixed, anchored learnable, free learnable, and homogeneous experts are compared under matched expert-head capacity and router architecture; learnable coefficient variants add only $3E$ scalar logits.

\subsubsection{Interaction-Conditioned Routing with Dense Distributions}
For a training instance $(u,i)$, MAGNET routes the interaction to a small subset of the structured experts in \Cref{sec:magnet-triplet-experts}.
\rev{The router conditions on the same three evidence sources available to the experts. We concatenate behavioral context with the history-induced user cues and the item's appearance and semantic cues:}
\begin{equation}
\mathbf q_{ui}=[\operatorname{Norm}(\mathbf r_u^B);\operatorname{Norm}(\mathbf r_i^B);\operatorname{Norm}(\mathbf h_u^{A});\operatorname{Norm}(\mathbf h_i^{A});\operatorname{Norm}(\mathbf h_u^{S});\operatorname{Norm}(\mathbf h_i^{S})]\in\mathbb{R}^{6d}.
\label{eq:router_query}
\end{equation}
\rev{Here, $\operatorname{Norm}(\cdot)$ denotes slot-wise $\ell_2$ normalization applied independently to each of the six $d$-dimensional query blocks before concatenation. In contrast, $\operatorname{Cal}_m(\cdot)$ in \Cref{eq:cue-calibration} denotes source calibration applied to the projected source representations before template fusion. The query normalization prevents a high-norm source from dominating the gate by scale alone.} \rev{For the behavior-only control in \Cref{sec:template-design-analysis}, the appearance and semantic query slots are zero-masked while retaining the same $6d\!\rightarrow\!E$ gate:
$\mathbf q^{B\text{-only}}_{ui}=[\operatorname{Norm}(\mathbf r_u^B);\operatorname{Norm}(\mathbf r_i^B);\mathbf 0_d;\mathbf 0_d;\mathbf 0_d;\mathbf 0_d]\in\mathbb R^{6d}$.}

Our progressive routing scheme follows standard sparse MoE training while adding an entropy-triggered regularization schedule that first promotes broad population-level coverage and then sharpens instance-level assignments while retaining a coverage guard.

Specifically, $\mathbf q_{ui}$ is defined in \eqref{eq:router_query}
and produces a \emph{dense} routing distribution over all $E$ experts:
\begin{equation}
\boldsymbol{\pi}_{ui} = \mathrm{Softmax}(\mathrm{Router}(\mathbf{q}_{ui};\Theta_{\mathrm{rt}}))\in\mathbb{R}^{E}.
\end{equation}
We keep $\boldsymbol{\pi}_{ui}$ as dense probabilities for (i) computing entropy-based statistics and
(ii) applying stage-wise routing regularization, while the forward aggregation still uses sparse Top-$K$
experts (described in \Cref{sec:expert_activation}). Over a mini-batch $\mathcal{B}$, we aggregate $\{\boldsymbol{\pi}_{ui}\}_{(u,i)\in\mathcal{B}}$
to estimate routing entropy and trigger stage switching in \Cref{sec:entropy}.

\subsubsection{Sparse Expert Activation, Aggregation, and Scoring}
\label{sec:expert_activation}
For efficient computation, we activate the Top-$K$ experts, renormalize their
weights, and aggregate their scalar scores from \Cref{eq:expert-pair}:
\begin{equation}
\label{eq:sparse-routing}
\Gamma_{ui}=\operatorname{TopK}(\boldsymbol\pi_{ui},K),\qquad
\tilde{\pi}^{(e)}_{ui}=\frac{\pi^{(e)}_{ui}}{\sum_{e'\in\Gamma_{ui}}\pi^{(e')}_{ui}},\qquad
\hat y_{ui}=\sum_{e\in\Gamma_{ui}}
\tilde{\pi}^{(e)}_{ui}\,s^{(e)}_{ui}.
\end{equation}
In all experiments we use $\tilde{\boldsymbol\pi}_{ui}$ for aggregation; the dense $\boldsymbol\pi_{ui}$ is retained for entropy statistics and progressive routing regularization (\Cref{sec:entropy}).
We adopt standard sparse MoE training: gradients are back-propagated through the selected experts and their corresponding router weights.

\paragraph{Implementation of the learned blocks.}
For each modality $m\in\{A,S\}$, $f_m$ is an affine projection from the input
feature dimension to $d$; source calibration is then applied as defined in
\Cref{eq:cue-calibration}. The behavioral representation uses one shared
$d\!\rightarrow\!d$ affine projection before calibration. The router is an
affine gate $6d\!\rightarrow\!E$ followed by softmax. \rev{Each
$\phi(\cdot;\Theta_e)$ is an independently parameterized scalar MLP with
dimensions $2d\!\rightarrow\!d\!\rightarrow\!1$, using ReLU and dropout after
the hidden layer. Thus, the evidence projections and router are shared, while
the lightweight scalar expert heads are branch-specific.}

\subsection{Entropy-Triggered Progressive Routing for Training}
\label{sec:entropy}

\begin{algorithm}[t]
\caption{Progressive Routing Training of MAGNET}
\label{alg:magnet_train}
\resizebox{0.9\columnwidth}{!}{%
\begin{minipage}{\columnwidth}
\SetKwInOut{KwRequire}{Require}
\SetKwInOut{KwEnsure}{Ensure}

\KwRequire{Observed edges $\mathcal{E}$; graphs $G_{UI}$ and $G_{UIG}$ (if $\mathcal{E}^+\neq\emptyset$);
mini-batch $\mathcal{B}\subset\mathcal{E}$ with $b:=|\mathcal{B}|$, neg. ratio $\rho$; experts $(E,K)$;
entropy $(H^*,W)$; weights $(\lambda,\lambda_{\mathrm{ctr}},\lambda_r,\lambda_{\rm anchor})$; temperature $\tau$; maximum update budget $T$; params $\Theta$}
\KwEnsure{Trained parameters $\Theta$}

$\textit{stage}\leftarrow 1;\; n\leftarrow 0$ \tcp*[r]{counter for persistent coverage}
$T_{\rm exp}\leftarrow\lfloor0.05T\rfloor$ \tcp*[r]{$T$: preset maximum update budget}

\For{$t\leftarrow 1$ \KwTo $T$}{

  Sample $\mathcal{B}\subset\mathcal{E}$ with $|\mathcal{B}|=b$ \tcp*[r]{(I) mini-batch sampling on observed edges}
  For each $(u,i)\in\mathcal{B}$, sample negatives $\mathcal{J}_u\sim \textsc{Neg}(u;\rho)$ with $j\notin \mathcal{I}_u$\;

  $\mathcal{L}_{\mathrm{BPR}}\leftarrow 0;\; \Pi\leftarrow \emptyset$ \tcp*[r]{\rev{(II) BPR loss + collect positive-pair dense routings}}
  \ForEach{$(u,i)\in\mathcal{B}$}{
$(\hat{y}_{ui},\boldsymbol{\pi}_{ui}) \leftarrow \textsc{MAGNET-Score}(u,i;\Theta)$\;
    $\Pi \leftarrow \Pi \cup \{\boldsymbol{\pi}_{ui}\}$\;
    $\{\hat{y}_{uj}\}_{j\in\mathcal{J}_u}\leftarrow \textsc{MAGNET-Score}(u,\mathcal{J}_u;\Theta).\hat{y}$ \tcp*[r]{\rev{score only; negative routings excluded from $\Pi$}}
    \rev{$\mathcal{L}_{\mathrm{BPR}} \mathrel{+}= -(b\rho)^{-1}\sum_{j\in\mathcal{J}_u}\log\sigma(\hat{y}_{ui}-\hat{y}_{uj})$}
  }

  $\mathcal{L}_{\mathrm{ctr}}\leftarrow 0$ \tcp*[r]{(III) optional dual-view alignment}
  \If{dual-view encoding is enabled \textbf{and} $\lambda_{\mathrm{ctr}}>0$}{
    $\mathcal{L}_{\mathrm{ctr}}\leftarrow \textsc{InfoNCE}(\cdot;\tau)$ \tcp*[r]{align $z^{UI}$ vs $z^{UIG}$  \Cref{sec:objective_routing_reg}}
  }

  \rev{$\bar{\boldsymbol{\pi}}\leftarrow \frac1{|\Pi|}\sum_{\boldsymbol{\pi}\in\Pi}\boldsymbol{\pi}$; $H_{\rm cov}\leftarrow \mathrm{Ent}(\bar{\boldsymbol{\pi}})/\log E$}\;
  \rev{$H_{\rm inst}\leftarrow \frac1{|\Pi|}\sum_{\boldsymbol{\pi}\in\Pi}\mathrm{Ent}(\boldsymbol{\pi})/\log E$} \tcp*[r]{coverage vs. decisiveness}
  \rev{$\omega(t)\leftarrow \mathbb I[t\le T_{\rm exp}]$; $\mathcal{L}_{\rm route}\leftarrow \lambda_{\rm cov}(1-H_{\rm cov})+\lambda_{\rm exp}\omega(t)(1-H_{\rm inst})$} \tcp*[r]{Stage 1}
  \If{$\textit{stage}=2$}{\rev{$\mathcal{L}_{\rm route}\leftarrow \lambda_{\rm MI}(H_{\rm inst}-H_{\rm cov})+\lambda_{\rm guard}(1-H_{\rm cov})$}}

  $\mathcal{L}\leftarrow \mathcal{L}_{\mathrm{BPR}}+\lambda_{\mathrm{ctr}}\mathcal{L}_{\mathrm{ctr}}
  +\mathcal{L}_{\rm route}
  +\lambda_{\rm anchor}\mathcal{L}_{\rm anchor}
  +\lambda\|\Theta\|_2^2$\;
  Backpropagate $\mathcal L$ and update $\Theta$ with Adam\;
  \If{$H_{\rm cov}\ge H^*$}{ $n\leftarrow n+1$ }\Else{ $n\leftarrow 0$ }
  \If(\tcp*[f]{effective at step $t+1$}){$\textit{stage}=1$ \textbf{and} $n\ge W$}{ $\textit{stage}\leftarrow 2$ \tcp*[r]{one-way switch} }
}
\KwRet{$\Theta$}\;
\end{minipage}}
\end{algorithm}

\subsubsection{Routing Entropy and Effective Experts}
\label{sec:routing_entropy}
\Cref{alg:magnet_train} summarizes entropy-triggered progressive routing; we describe its key components below.

Given a mini-batch $\mathcal{B}$, we define the batch-mean routing distribution
\begin{equation}
\bar{\boldsymbol{\pi}}=\frac{1}{|\mathcal{B}|}\sum_{(u,i)\in\mathcal{B}}\boldsymbol{\pi}_{ui},
\label{eq:pi_bar}
\end{equation}
and define the coverage, instance-level, and derived routing statistics jointly:
\begin{equation}
\begin{aligned}
\rev{H_{\mathrm{cov}}}&\rev{=\frac{H(\bar{\boldsymbol{\pi}})}{\log E}},
&\rev{H_{\mathrm{inst}}}&\rev{=\frac{1}{|\mathcal B|\log E}\sum_{(u,i)\in\mathcal B}H(\boldsymbol\pi_{ui})},\\
\rev{I_{\mathrm{route}}}&\rev{=H_{\mathrm{cov}}-H_{\mathrm{inst}}},
&N_{\mathrm{eff}}&=\exp\!\big(H(\bar{\boldsymbol{\pi}})\big),
\end{aligned}
\label{eq:two-entropies}
\end{equation}
where $H(\mathbf p)=-\sum_e p(e)\log p(e)$. $H_{\mathrm{cov}}$ measures population-level coverage of the dense pre-Top-$K$ routing distribution, whereas $H_{\mathrm{inst}}$ measures uncertainty within individual dense assignments. Their difference $I_{\mathrm{route}}$ is large when assignments are individually decisive and collectively diverse, and $N_{\mathrm{eff}}$ is the effective support size of the dense batch-mean routing distribution. These statistics describe routing mass before sparsification; the forward path activates only the selected set $\Gamma_{ui}$ in \Cref{eq:sparse-routing}.
\rev{Here $\mathcal B$ contains observed positive edges. The routing statistics aggregate their dense distributions $\boldsymbol\pi_{ui}$; sampled negatives contribute to $\mathcal L_{\mathrm{BPR}}$ but not to $H_{\mathrm{cov}}$ or $H_{\mathrm{inst}}$.}

\subsubsection{Entropy-Triggered Two-Stage Schedule}
\label{sec:two_stage_schedule}

We adopt an entropy-triggered two-stage schedule to progressively transition from
\emph{broad routing-mass coverage} to \emph{confident specialization} in MoE routing.
Intuitively, early in training the router is unstable and its dense probability mass may collapse onto a few experts; thus we first
encourage broad population-level routing-mass coverage. Once this mass becomes sufficiently dispersed, we switch to a
second stage that sharpens and stabilizes routing for more consistent expert specialization.

Concretely, at each optimization step $t$ we compute $H_{\mathrm{cov}}$ in
\Cref{eq:two-entropies}. The current step uses the stage active at its start; after that step's loss is formed, its coverage statistic updates the persistence counter, and any resulting transition takes effect at step $t+1$:
\begin{equation}
n_t=
\begin{cases}
n_{t-1}+1, & \text{if }H_{\mathrm{cov}}\ge H^*,\\
0,   & \text{otherwise},
\end{cases}
\qquad
\textit{stage}_{t+1}=
\begin{cases}
2, & \text{if } \textit{stage}_{t}=1\ \text{and}\ n_t\ge W,\\
\textit{stage}_{t}, & \text{otherwise},
\end{cases}
\label{eq:stage_switch}
\end{equation}
where $H^*\in[0,1]$ is an entropy threshold and $W=4$ is a persistence window counted in optimization steps. The transition from Stage~1 to Stage~2 is one-way.

\noindent\textbf{Stage semantics.}
\rev{Stage~1 emphasizes broad population-level routing-mass coverage and avoids premature routing-mass collapse. Stage~2 reduces conditional assignment entropy while preserving population-level routing coverage. The transition is triggered by $H_{\mathrm{cov}}$ according to \Cref{eq:stage_switch}; $H_{\mathrm{inst}}$ separately characterizes per-instance decisiveness.}

\subsubsection{Objective with Stage-Specific Routing Regularization}
\label{sec:objective_routing_reg}

For preference learning, we average the standard pairwise objective over sampled pairs,
$\mathcal L_{\mathrm{BPR}}=N_{\mathrm{pair}}^{-1}\sum_{(u,i)\in\mathcal B}\sum_{j\in\mathcal J_u}-\log\sigma(\hat y_{ui}-\hat y_{uj})$,
where $N_{\mathrm{pair}}=\sum_{(u,i)\in\mathcal B}|\mathcal J_u|$, $(u,i)\in\mathcal D_{tr}$, and $j\notin\mathcal I_u$.

For the dual-view setting, we optionally align the two structural representations with symmetric InfoNCE. For node type $X\in\{U,I\}$ and views $a,b\in\{UI,UIG\}$, let
\begin{equation}
\begin{aligned}
\ell_X^{a\rightarrow b}
&=-\frac{1}{|X_{\mathcal{B}}|}\sum_{x\in X_{\mathcal{B}}}
\log\frac{\exp(\operatorname{sim}(\mathbf{z}_x^a,\mathbf{z}_x^b)/\tau)}
{\sum_{x'\in X_{\mathcal{B}}}\exp(\operatorname{sim}(\mathbf{z}_x^a,\mathbf{z}_{x'}^b)/\tau)},\\
\mathcal L_{\mathrm{ctr}}
&=\frac{1}{2}\sum_{X\in\{U,I\}}\left(\ell_X^{UI\rightarrow UIG}+\ell_X^{UIG\rightarrow UI}\right),
\end{aligned}
\label{eq:ctr}
\end{equation}
where $X_{\mathcal{B}}$ contains the users or items appearing in mini-batch $\mathcal{B}$, and $\operatorname{sim}$ is cosine similarity. For the reported MAGNET-DV results, we use $\tau=0.5$ and $\lambda_{\mathrm{ctr}}=0.01$, except in the explicit view-contrastive ablation. The contrastive objective is disabled for single-graph configurations, including the observed-only, unfiltered-overlay, and filtered-overlay controls, for which no independently encoded second structural view is available.

\noindent\textbf{Stage-specific routing objectives.}
We regularize the dense distribution $\boldsymbol\pi_{ui}$ using the two quantities in \Cref{eq:two-entropies}. During Stage~1, we minimize the coverage and exploration deficits:
\begin{equation}
\mathcal L_{\mathrm{route}}^{(1)}
=\lambda_{\mathrm{cov}}(1-H_{\mathrm{cov}})
+\lambda_{\mathrm{exp}}\,\omega(t)(1-H_{\mathrm{inst}}),
\label{eq:stage1-route}
\end{equation}
where \rev{$T$ is the maximum optimization-update budget, $T_{\mathrm{exp}}=\lfloor0.05T\rfloor$, and $\omega(t)=\mathbb I[t\le T_{\mathrm{exp}}]$. The transient exploration term is active during these initial updates}, while stage switching remains governed by \Cref{eq:stage_switch}; the same rule is used across datasets. The first term prevents population-level routing-mass collapse, and the transient second term avoids making the earliest assignments overconfident.

After the coverage trigger is sustained for $W$ steps, Stage~2 maximizes the routing mutual-information proxy while retaining an explicit coverage-preservation term:
\begin{equation}
\rev{\mathcal L_{\mathrm{route}}^{(2)}
=\lambda_{\mathrm{MI}}(H_{\mathrm{inst}}-H_{\mathrm{cov}})
+\lambda_{\mathrm{guard}}(1-H_{\mathrm{cov}}).}
\label{eq:stage2-route}
\end{equation}
\rev{The first term encourages assignments that are individually sharp yet collectively diverse; the second acts as a continuous coverage guard throughout Stage~2. This objective separates per-instance decisiveness from batch-level routing-mass coverage.}

Putting everything together, the overall objective is
\begin{equation}
\mathcal{L}=\mathcal{L}_{\mathrm{BPR}}
+\lambda_{\mathrm{ctr}}\mathcal{L}_{\mathrm{ctr}}
+\mathbb I[\mathrm{stage}=1]\mathcal L_{\mathrm{route}}^{(1)}
+\mathbb I[\mathrm{stage}=2]\mathcal L_{\mathrm{route}}^{(2)}
+\lambda_{\mathrm{anchor}}\mathcal L_{\mathrm{anchor}}
+\lambda\|\Theta\|_2^2,
\label{eq:overall_obj}
\end{equation}
where $\Theta$ collects all trainable parameters. $\lambda_{\mathrm{anchor}}=0$ for fixed templates and is set to $0.05$ for the anchored learnable control on every dataset.

\section{Experiments}
\label{sec:experiments}

Our experiments answer five linked questions: \textbf{RQ1} whether multimodal evidence usefulness varies across interactions and fixed fusion rules exhibit regime-dependent limitations; \textbf{RQ2} whether MAGNET is competitive under matched protocols, including non-Amazon, tail-item, and low-history warm-start user settings; \textbf{RQ3} whether structure-aware evidence preparation through separate structural views is useful; \textbf{RQ4} whether structured experts and modality-aware routing outperform matched alternatives; and \textbf{RQ5} \rev{whether the two-entropy schedule maintains broad routing coverage during instance-level specialization without collapse}.

\subsection{Datasets and Preprocessing}
\label{sec:datasets-prep}
We evaluate on four public Amazon review datasets---Baby, Sports, Clothing, and Electronics---\rev{and on MicroLens-100K~\cite{Ni2023MicroLens}, a non-commerce short-video benchmark}. The Amazon subsets vary substantially in catalog size and density, while \rev{MicroLens-100K extends the evaluation to a short-video domain with a distinct interaction distribution and content regime}.

For Amazon, we use the public MMRec data package, including fixed interactions, \rev{4,096-dimensional CNN visual features and 384-dimensional Sentence-Transformer textual features}.\footnote{\url{https://drive.google.com/drive/folders/13cBy1EA_saTUuXxVllKgtfci2A09jyaG}} \rev{For MicroLens-100K, we use its public interaction file and extracted cover-image/title features distributed through the MMRec-compatible release. Within each dataset, every baseline and MAGNET use the same interaction split and feature tensors. The source-specific diagnostic analysis is therefore conditioned on this shared feature setting.} Basic statistics are summarized in \Cref{tab:dataset_stats}.

\begin{table}[H]
\centering
\small
\caption{Statistics of the five evaluation datasets. All counts are computed before the fixed train/validation/test split.}
\label{tab:dataset_stats}
\setlength{\tabcolsep}{4.5pt}
\begin{tabular}{llrrrrr}
\toprule
Dataset & Domain & Users & Items & Interactions & Density & Avg./user\\
\midrule
\multicolumn{7}{l}{\textbf{Amazon Review domains}}\\[-1pt]
Baby        & E-commerce & 19,445  & 7,050  & 160,792   & 0.1173\% & 8.27\\
Sports      & E-commerce & 35,598  & 18,357 & 296,337   & 0.0453\% & 8.32\\
Clothing    & E-commerce & 39,387  & 23,033 & 278,677   & 0.0307\% & 7.08\\
Electronics & E-commerce & 192,403 & 63,001 & 1,689,188 & 0.0139\% & 8.78\\
\addlinespace[1pt]
\cmidrule(lr){1-7}
\multicolumn{7}{l}{\rev{\textbf{Non-Amazon benchmark}}}\\[-1pt]
\rev{MicroLens-100K~\cite{Ni2023MicroLens}}
            & \rev{Short video} & \rev{100,000} & \rev{19,738} & \rev{719,405} & \rev{0.0364\%} & \rev{7.19}\\
\bottomrule
\end{tabular}
\end{table}

We treat observed user--item interactions in $\mathbf{R}$ as positive implicit feedback. For the four Amazon datasets, we directly use the processed interaction files released with MMRec, which satisfy the requirement of at least four interactions per user; no additional user filtering is performed during MAGNET data loading. For MicroLens-100K, preprocessing retains users with at least four interactions.
To avoid information leakage, all user-history-dependent constructions, including modality-profile construction, the augmentation set $\mathcal{C}_{\mathrm{aug}}(u)$, and induced edges $\mathcal{E}^{+}$, use only the training history $\mathcal{I}_u$.

\rev{We retain the fixed $8{:}1{:}1$ train/validation/test partition supplied by the shared preprocessing pipeline without re-splitting. Every main-table method is trained using the same five optimization seeds in matched repeated trials: $9$, $672$, $5368$, $12784$, and $2025$. We report the arithmetic mean and sample standard deviation (denominator $n-1$) across these matched runs.}

\subsection{Baselines and Implementation Sources}
\label{sec:baselines}
We compare MAGNET against competitive baselines that progressively incorporate multimodal signals, from interaction-only collaborative filtering, through dual-graph/structure-learning multimodal methods, to recent contrastive/self-supervised multimodal models:

\begin{itemize}[leftmargin=*,itemsep=1pt,topsep=1pt]
    \item \textbf{Interaction-only recommenders:} BPR~\cite{Rendle2009BPR}, LightGCN~\cite{He2020LightGCN}.
    \item \textbf{Graph-structure and fusion-based multimodal recommenders:} LATTICE~\cite{Zhang00WWW21}, FREEDOM~\cite{ZhouS23}, and \rev{COHESION~\cite{Xu2025COHESION}}.
    \item \textbf{Contrastive and self-supervised multimodal recommenders:} MGCN~\cite{Yu2023MM}, GUME~\cite{Lin2024GUME}, SOIL~\cite{Su2024SOIL}, SMORE~\cite{Ong2025Spectrum}, MIG-GT~\cite{Hu2025Modality}, MENTOR~\cite{Xu2025MENTOR}, CMDL~\cite{Lin2025Contrastive}, and ITCoHD-MRec~\cite{Hao2025ITCoHD}.
\end{itemize}

The Amazon experiments include all listed methods with compatible implementations and feature inputs. \rev{For MicroLens-100K, we evaluate LightGCN, MMGCN~\cite{Wei2019MMGCN}, GRCN~\cite{Wei2020Graph}, BM3~\cite{Zhou2023Bootstrap}, FREEDOM, GUME, and MGCN under the same dataset-level protocol.} We use official implementations when available and verified re-implementations otherwise. All baselines retain their method-native objectives.

\rev{A VI-MMRec-style filtered overlay~\cite{Xu2026VIMMRec} is evaluated as a structural-augmentation control in \Cref{sec:vi_comparison}.}

Validation-selected hyperparameters follow the early-stopping rule in \Cref{sec:protocol}, while globally shared design settings are fixed as summarized in \Cref{tab:hyper}. Code sources, dataset-specific configurations, and reproduction status are documented with the released implementation.

\rev{The released repository provides the MAGNET implementation, graph-construction pipeline, dataset configurations, repeated-run scripts, evaluation utilities, and a reproduction manifest.}

\subsection{Training and Evaluation Protocol}
\label{sec:protocol}

\subsubsection{Training Setup and Reproducibility}
The dataset-level preprocessing, fixed split, shared features, and repeated-seed design are defined once in \Cref{sec:datasets-prep}. \rev{Unless otherwise stated, diagnostic analyses use seed 9, one of the five prespecified repeated-run seeds. Statistical tests are reported only for repeated-run comparisons in the main tables.}

We retain each baseline's official training objective and method-specific sampling strategy; for example, BM3 uses its native bootstrap/self-supervised objective rather than pairwise negative sampling. MAGNET and the BPR-style controls use mini-batches of 1024 with uniform 1:1 positive-to-negative sampling. We use Adam for optimization and apply early stopping on the validation set
with a patience of 5 and a maximum of 200 epochs. For each run, the checkpoint is selected by the best
validation performance (by default, NDCG@20).

Schedule variables $t$, $T$, $T_{\rm exp}$, and $W$ are measured in optimizer updates, whereas $T_{\max}$ denotes the maximum epoch budget.

\subsubsection{Evaluation Protocol}
We evaluate all methods with Recall@$N$ and NDCG@$N$, where $N \in \{10,20\}$, using full-sort ranking over the complete item catalog. For each user, items observed in the training history are masked before ranking; the held-out validation or test items provide the relevant set for that split. During test evaluation we do not additionally mask the held-out validation item, so it remains an ordinary non-relevant candidate under the same protocol for every method. Scores are computed in item chunks only to control memory use, which does not alter the candidate set or ranking. This protocol is shared by every method within a dataset. \Cref{sec:diagnostic-protocol} defines the separate full-sort candidate set used for diagnostic stratification. Unless otherwise specified, we use validation NDCG@20 for checkpoint selection and report test Recall@20 and NDCG@20 as the primary metrics.

\subsubsection{Hardware and Measurement}
All experiments are conducted on a single NVIDIA RTX 4090 GPU with 24\,GB memory, using a machine equipped
with a 16-core Intel Xeon(R) Platinum 8358P CPU and 120\,GB RAM. Training time is measured as wall-clock
seconds per epoch, averaged over the training trajectory after a short warm-up to avoid one-time initialization
effects. Peak GPU memory is reported as the maximum allocated GPU memory observed during training. \rev{These wall-clock measurements exclude one-time feature extraction and graph construction or materialization.}

\subsection{Hyperparameter Configuration}
\label{sec:hparams}

\Cref{tab:hyper} separates shared design settings from validation-tuned optimization parameters. We use one common architecture and one shared set of template, routing, and graph settings across all five datasets. The four routing coefficients share the fixed ratio below, so routing strength is controlled through the scalar $\lambda_r$.

The four routing coefficients are parameterized by one global strength:
\begin{equation}
(\lambda_{\rm exp},\lambda_{\rm cov},\lambda_{\rm MI},\lambda_{\rm guard})=\lambda_r(1,6,6,2),
\label{eq:routing_ratio}
\end{equation}
where the shared ratio emphasizes \rev{coverage promotion and coverage-preserving specialization} over the transient exploration term. \rev{We use $\lambda_r=0.30$ as a single shared operating point for every dataset and repeated run, without test-metric selection or per-dataset retuning.} The transient exploration window is $T_{\rm exp}=\lfloor0.05T\rfloor$; \Cref{sec:routing_sensitivity} examines the shared routing setting.

\begin{table}[!htbp]
\centering
\caption{Hyperparameter organization and final settings. \rev{Shared design values are used on all five datasets; standard optimizer controls are selected only on validation data.}}
\label{tab:hyper}
\small
\setlength{\tabcolsep}{4.5pt}
\renewcommand{\arraystretch}{1.08}
\resizebox{\linewidth}{!}{%
\begin{tabular}{llll}
\toprule
Group & Hyperparameters & Search/status & Final setting\\
\midrule
Architecture & $d,L,E,K$ & shared & $64,2,9,4$\\
Canonical templates & $\alpha,\beta,\delta,\xi$ & shared & $0.6,0.2,0.5,0.10$\\
Graph expansion & $k,r$ (neighbor size, retention budget) & shared & $20,200$\\
\addlinespace[1.5pt]
\rev{Stage trigger} & \rev{$(H^*,W)$} & \rev{shared} & \rev{$0.90,4$}\\
\rev{Transient exploration weighting} & \rev{$T_{\rm exp}$} & \rev{shared} & \rev{$\lfloor0.05T\rfloor$ updates}\\
\rev{Routing objective} & \rev{$\lambda_r$} & \rev{$\{0.05,0.10,0.20,0.30,0.40\}$ (shared search)} & \rev{$\mathbf{0.30}$}\\
\rev{Routing ratios} & \rev{$\lambda_{\rm exp}:\lambda_{\rm cov}:\lambda_{\rm MI}:\lambda_{\rm guard}$} & \rev{shared} & \rev{$1:6:6:2$}\\
\rev{Anchored-template prior} & \rev{$\lambda_{\rm anchor}$} & \rev{anchored control only} & \rev{$0.05$}\\
\addlinespace[1.5pt]
Optimization & $\eta,p_d,\lambda$ & standard validation tuning & $10^{-3},0.2,2\!\times\!10^{-5}$\\
View contrast & $\tau,\lambda_{\rm ctr}$ & shared & $0.5,0.01$\\
Training & batch, negatives, patience & shared & $1024,1{:}1,5$\\
\addlinespace[1.5pt]
\cmidrule(lr){1-4}
\rev{Diagnostic only} & \rev{$\tau_{\rm probe},\kappa$} & \rev{shared preset} & \rev{$0.10,0.45$}\\
\bottomrule
\end{tabular}
}
\end{table}

\FloatBarrier

\subsection{Motivating Diagnostic Protocol}
\label{sec:diagnostic-protocol}
All diagnostic results in \Cref{fig:motivation}(b--c) use seed~9 and the fixed test split. \Cref{fig:motivation} uses three diagnostic probes trained independently of MAGNET, using only behavior ($B$), appearance ($A$), or semantics ($S$), respectively. The probes share the same hidden dimension, pairwise ranking objective, scoring head, optimizer family, and candidate protocol; they differ only in the input source. For every held-out positive interaction $(u,i)$, all probes rank the target within the same full-sort diagnostic candidate set $\mathcal C_{\mathrm{diag}}(u)=\mathcal I\setminus \mathcal I_u$, i.e., the complete item catalog after masking the training-only history. No sampled negatives or popularity restriction are used. During test evaluation, the held-out validation item remains an ordinary non-relevant candidate in this diagnostic candidate set under the evaluation convention above. We convert each target rank to a normalized retrieval utility:
\begin{equation}
q_m(u,i)=1-\frac{\operatorname{rank}_m(i)-1}{|\mathcal C_{\mathrm{diag}}(u)|-1},
\qquad m\in\{B,A,S\},
\label{eq:diagnostic-utility}
\end{equation}
and normalize the three utilities into source shares $p_m(u,i)$ with a fixed-temperature softmax using $\tau_{\rm probe}{=}0.10$. We use a shared preset dominance threshold $\kappa{=}0.45$ across all datasets. An interaction is assigned to source $m$ when $p_m$ is the largest share and exceeds $\kappa$; otherwise it is assigned to the \emph{Mixed} regime. ``Mixed'' identifies interactions without a clearly dominant probe and need not entail rank disagreement.

To test whether one fixed fusion rule suffices, we independently train four capacity-matched models with fixed source coefficients: $B$-heavy $(0.6,0.2,0.2)$, $A$-heavy $(0.2,0.6,0.2)$, $S$-heavy $(0.2,0.2,0.6)$, and Balanced $(1/3,1/3,1/3)$. For \Cref{fig:motivation}(c), we aggregate diagnostic interactions at the interaction level across all five datasets. Let $\mathcal{D}_{d,r}$ denote the test interactions from dataset $d$ assigned to regime $r$. For fixed fusion rule $j$, we compute
\begin{equation}
s_{r,j}=\frac{\sum_d\sum_{(u,i)\in\mathcal{D}_{d,r}}\mathrm{NDCG@20}_{j}(u,i)}{\sum_d|\mathcal{D}_{d,r}|}.
\label{eq:diagnostic-pooling}
\end{equation} \Cref{fig:motivation}(c) reports the row-normalized score $\widetilde{s}_{r,j}=100\,s_{r,j}/\max_{j'}s_{r,j'}$. Thus, 100 identifies the best fixed rule within a stratum, whereas 92.4 means that the corresponding rule attains 92.4\% of that stratum's best score. Across strata, the winning rule changes, yielding the crossover pattern in \Cref{fig:motivation}(c). \rev{Probe assignments are used only to form post-hoc diagnostic strata; MAGNET and the fixed-fusion controls are trained and evaluated independently.} All strategies use the same diagnostic candidate set. We additionally examine $\kappa\in\{0.40,0.45,0.50\}$. Pairwise Spearman correlations among the 20-dimensional vectors of unrounded dataset-by-stratum proportions are 0.977, 0.970, and 0.944 for $0.40$ vs.\ $0.45$, $0.45$ vs.\ $0.50$, and $0.40$ vs.\ $0.50$, respectively. The identity of the four row-wise winning rules is unchanged.

\subsection{Overall Performance}
\label{sec:overall}

Tables~\ref{tab:main_results} and~\ref{tab:microlens_results} report the primary results on four Amazon domains and the non-Amazon MicroLens-100K benchmark. MAGNET-DV is the default model selected on validation data.

% Marked-revision layout for the Amazon main table.
% The full table is updated from the author-provided five-seed summaries.
\newcommand{\baseunc}[1]{{\color{red}\hspace{0.2pt}\scalebox{0.66}{$\pm #1$}}}
\newcommand{\basecell}[2]{#1\baseunc{#2}}
\newcommand{\basesecond}[2]{\underline{#1}\baseunc{#2}}
\newcommand{\revisioncell}[2]{{\color{red}#1\hspace{0.2pt}\scalebox{0.66}{$\pm #2$}}}
\newcommand{\revisionbest}[2]{{\color{red}\textbf{#1}\hspace{0.2pt}\scalebox{0.66}{$\pm #2$}}}
\newcommand{\revisionsecond}[2]{{\color{red}\underline{#1}\hspace{0.2pt}\scalebox{0.66}{$\pm #2$}}}
% Rank highlights are content-sized boxes rather than full cell fills.  The
% surrounding tabular padding therefore leaves a visible white gutter between
% adjacent highlighted values, matching the more refined original treatment.
\newcommand{\mainrankone}[1]{{\setlength{\fboxsep}{0.72pt}\colorbox{TableTopOne}{\strut #1}}}
\newcommand{\mainranktwo}[1]{{\setlength{\fboxsep}{0.72pt}\colorbox{TableTopTwo}{\strut #1}}}
\newcommand{\mainrankthree}[1]{{\setlength{\fboxsep}{0.72pt}\colorbox{TableTopThree}{\strut #1}}}

\begin{table}[!t]
\centering
\caption{Amazon results over five matched seeds on one fixed split \rev{(mean $\pm$ sample standard deviation)}. Pale coral, blue, and green backgrounds mark the first-, second-, and third-ranked values; boldface marks the overall best, and underlining marks the strongest non-MAGNET result. \rev{$^{*}p<0.05$ and $^{**}p<0.01$ under two-sided one-sample $t$-tests of five seed-wise performance differences ($n=5$, $df=4$), without family-wise correction.}}
\label{tab:main_results}
\fontsize{6.4}{7.0}\selectfont
\setlength{\tabcolsep}{1.45pt}
\renewcommand{\arraystretch}{0.98}
\resizebox{0.995\linewidth}{!}{%
\begin{tabular}{lcccccccc}
\toprule
\multirow{2}{*}{Method} & \multicolumn{4}{c}{Baby} & \multicolumn{4}{c}{Sports}\\
\cmidrule(lr){2-5}\cmidrule(lr){6-9}
& R@10 & R@20 & N@10 & N@20 & R@10 & R@20 & N@10 & N@20\\
\midrule
\rowcolor{TableGroup}\multicolumn{9}{l}{\textit{General recommender baselines}}\\[-1pt]
BPR       & \basecell{0.0357}{0.0025} & \basecell{0.0575}{0.0027} & \basecell{0.0192}{0.0012} & \basecell{0.0249}{0.0012} & \basecell{0.0432}{0.0025} & \basecell{0.0653}{0.0026} & \basecell{0.0241}{0.0015} & \basecell{0.0298}{0.0018}\\
LightGCN  & \basecell{0.0478}{0.0021} & \basecell{0.0753}{0.0024} & \basecell{0.0257}{0.0013} & \basecell{0.0328}{0.0015} & \basecell{0.0569}{0.0026} & \basecell{0.0863}{0.0027} & \basecell{0.0311}{0.0011} & \basecell{0.0387}{0.0013}\\
\addlinespace[1pt]
\rowcolor{TableGroup}\multicolumn{9}{l}{\textit{Multimodal recommender baselines}}\\[-1pt]
LATTICE   & \basecell{0.0533}{0.0031} & \basecell{0.0853}{0.0034} & \basecell{0.0285}{0.0015} & \basecell{0.0368}{0.0017} & \basecell{0.0618}{0.0037} & \basecell{0.0949}{0.0037} & \basecell{0.0337}{0.0018} & \basecell{0.0422}{0.0019}\\
FREEDOM   & \basecell{0.0627}{0.0016} & \basecell{0.0992}{0.0017} & \basecell{0.0330}{0.0008} & \basecell{0.0424}{0.0010} & \basecell{0.0714}{0.0021} & \basecell{0.1086}{0.0023} & \basecell{0.0384}{0.0014} & \basecell{0.0479}{0.0015}\\
MGCN      & \basecell{0.0621}{0.0024} & \basecell{0.0966}{0.0025} & \basecell{0.0340}{0.0013} & \basecell{0.0428}{0.0015} & \basecell{0.0729}{0.0026} & \basecell{0.1107}{0.0030} & \basecell{0.0397}{0.0015} & \basecell{0.0496}{0.0017}\\
GUME      & \basecell{0.0674}{0.0016} & \basecell{0.1045}{0.0018} & \mainrankthree{\basesecond{0.0366}{0.0009}} & \mainrankthree{\basesecond{0.0462}{0.0010}} & \basecell{0.0775}{0.0014} & \mainrankthree{\basesecond{0.1160}{0.0016}} & \basecell{0.0425}{0.0009} & \basecell{0.0525}{0.0011}\\
SOIL      & \basecell{0.0680}{0.0019} & \basecell{0.1027}{0.0022} & \basecell{0.0364}{0.0011} & \basecell{0.0454}{0.0011} & \mainrankthree{\basesecond{0.0786}{0.0025}} & \basecell{0.1156}{0.0025} & \mainrankthree{\basesecond{0.0435}{0.0013}} & \mainrankthree{\basesecond{0.0530}{0.0014}}\\
SMORE     & \basecell{0.0679}{0.0027} & \basecell{0.1034}{0.0028} & \basecell{0.0364}{0.0014} & \basecell{0.0456}{0.0015} & \basecell{0.0759}{0.0024} & \basecell{0.1139}{0.0026} & \basecell{0.0407}{0.0014} & \basecell{0.0504}{0.0015}\\
MIG-GT    & \basecell{0.0667}{0.0023} & \basecell{0.1025}{0.0027} & \basecell{0.0363}{0.0012} & \basecell{0.0453}{0.0015} & \basecell{0.0753}{0.0026} & \basecell{0.1128}{0.0029} & \basecell{0.0414}{0.0013} & \basecell{0.0511}{0.0015}\\
MENTOR    & \basecell{0.0676}{0.0018} & \basecell{0.1044}{0.0019} & \basecell{0.0361}{0.0007} & \basecell{0.0449}{0.0008} & \basecell{0.0764}{0.0024} & \basecell{0.1138}{0.0027} & \basecell{0.0410}{0.0013} & \basecell{0.0511}{0.0013}\\
CMDL      & \basecell{0.0650}{0.0023} & \basecell{0.0913}{0.0027} & \basecell{0.0314}{0.0012} & \basecell{0.0393}{0.0013} & \basecell{0.0726}{0.0025} & \basecell{0.1097}{0.0027} & \basecell{0.0391}{0.0013} & \basecell{0.0471}{0.0014}\\
ITCoHD-MRec & \basecell{0.0668}{0.0023} & \basecell{0.1016}{0.0025} & \basecell{0.0361}{0.0011} & \basecell{0.0452}{0.0012} & \basecell{0.0734}{0.0023} & \basecell{0.1103}{0.0027} & \basecell{0.0398}{0.0011} & \basecell{0.0492}{0.0013}\\
\textcolor{red}{COHESION} & \mainrankthree{\revisionsecond{0.0681}{0.0023}} & \mainrankthree{\revisionsecond{0.1052}{0.0026}} & \revisioncell{0.0364}{0.0012} & \revisioncell{0.0453}{0.0014} & \revisioncell{0.0757}{0.0028} & \revisioncell{0.1135}{0.0029} & \revisioncell{0.0409}{0.0016} & \revisioncell{0.0502}{0.0018}\\
\midrule
\rowcolor{TableGroup}\multicolumn{9}{l}{\textit{MAGNET variants}}\\[-1pt]
\textcolor{red}{MAGNET-SV} & \mainranktwo{\revisioncell{0.0698}{0.0016}} & \mainranktwo{\revisioncell{0.1068}{0.0018}} & \mainranktwo{\revisioncell{0.0375}{0.0010}} & \mainranktwo{\revisioncell{0.0475}{0.0011}} & \mainrankone{\revisionbest{0.0817}{0.0020}} & \mainranktwo{\revisioncell{0.1175}{0.0021}} & \mainranktwo{\revisioncell{0.0449}{0.0009}} & \mainranktwo{\revisioncell{0.0548}{0.0010}}\\
\textcolor{red}{\textbf{MAGNET-DV}} & \mainrankone{\revisionbest{0.0708\psig{*}}{0.0015}} & \mainrankone{\revisionbest{0.1086\psig{*}}{0.0018}} & \mainrankone{\revisionbest{0.0376\psig{*}}{0.0008}} & \mainrankone{\revisionbest{0.0481\psig{**}}{0.0009}} & \mainranktwo{\revisioncell{0.0808\psig{*}}{0.0017}} & \mainrankone{\revisionbest{0.1188\psig{**}}{0.0018}} & \mainrankone{\revisionbest{0.0455\psig{**}}{0.0011}} & \mainrankone{\revisionbest{0.0554\psig{**}}{0.0012}}\\
\bottomrule
\end{tabular}}
\par\vspace{4pt}
\resizebox{0.995\linewidth}{!}{%
\begin{tabular}{lcccccccc}
\toprule
\multirow{2}{*}{Method} & \multicolumn{4}{c}{Clothing} & \multicolumn{4}{c}{Electronics}\\
\cmidrule(lr){2-5}\cmidrule(lr){6-9}
& R@10 & R@20 & N@10 & N@20 & R@10 & R@20 & N@10 & N@20\\
\midrule
\rowcolor{TableGroup}\multicolumn{9}{l}{\textit{General recommender baselines}}\\[-1pt]
BPR       & \basecell{0.0206}{0.0013} & \basecell{0.0304}{0.0014} & \basecell{0.0114}{0.0008} & \basecell{0.0139}{0.0009} & \basecell{0.0347}{0.0019} & \basecell{0.0518}{0.0025} & \basecell{0.0197}{0.0009} & \basecell{0.0238}{0.0012}\\
LightGCN  & \basecell{0.0360}{0.0020} & \basecell{0.0543}{0.0020} & \basecell{0.0197}{0.0009} & \basecell{0.0242}{0.0010} & \basecell{0.0362}{0.0026} & \basecell{0.0538}{0.0027} & \basecell{0.0203}{0.0015} & \basecell{0.0249}{0.0017}\\
\addlinespace[1pt]
\rowcolor{TableGroup}\multicolumn{9}{l}{\textit{Multimodal recommender baselines}}\\[-1pt]
LATTICE   & \basecell{0.0459}{0.0029} & \basecell{0.0703}{0.0031} & \basecell{0.0254}{0.0014} & \basecell{0.0306}{0.0017} & \revisioncell{0.0410}{0.0035} & \revisioncell{0.0619}{0.0039} & \revisioncell{0.0231}{0.0015} & \revisioncell{0.0285}{0.0017}\\
FREEDOM   & \basecell{0.0629}{0.0024} & \basecell{0.0940}{0.0029} & \basecell{0.0341}{0.0014} & \basecell{0.0420}{0.0017} & \basecell{0.0382}{0.0029} & \basecell{0.0589}{0.0032} & \basecell{0.0209}{0.0017} & \basecell{0.0262}{0.0019}\\
MGCN      & \basecell{0.0642}{0.0020} & \basecell{0.0946}{0.0022} & \basecell{0.0347}{0.0012} & \basecell{0.0428}{0.0013} & \basecell{0.0441}{0.0024} & \basecell{0.0648}{0.0026} & \basecell{0.0245}{0.0011} & \basecell{0.0302}{0.0013}\\
GUME      & \mainrankthree{\basesecond{0.0702}{0.0018}} & \mainrankthree{\basesecond{0.1022}{0.0020}} & \mainrankthree{\basesecond{0.0383}{0.0010}} & \mainrankthree{\basesecond{0.0465}{0.0012}} & \mainrankthree{\basesecond{0.0460}{0.0022}} & \mainrankthree{\basesecond{0.0681}{0.0022}} & \mainrankthree{\basesecond{0.0254}{0.0012}} & \mainrankthree{\basesecond{0.0311}{0.0014}}\\
SOIL      & \basecell{0.0688}{0.0026} & \basecell{0.0999}{0.0030} & \basecell{0.0378}{0.0012} & \basecell{0.0457}{0.0014} & \basecell{0.0455}{0.0025} & \basecell{0.0677}{0.0029} & \basecell{0.0250}{0.0011} & \basecell{0.0305}{0.0012}\\
SMORE     & \basecell{0.0659}{0.0029} & \basecell{0.0986}{0.0032} & \basecell{0.0360}{0.0016} & \basecell{0.0443}{0.0019} & \revisioncell{0.0433}{0.0022} & \revisioncell{0.0648}{0.0024} & \revisioncell{0.0239}{0.0013} & \revisioncell{0.0295}{0.0015}\\
MIG-GT    & \basecell{0.0635}{0.0024} & \basecell{0.0931}{0.0029} & \basecell{0.0346}{0.0013} & \basecell{0.0422}{0.0016} & \revisioncell{0.0431}{0.0024} & \revisioncell{0.0643}{0.0028} & \revisioncell{0.0241}{0.0013} & \revisioncell{0.0297}{0.0015}\\
MENTOR    & \basecell{0.0667}{0.0023} & \basecell{0.0989}{0.0024} & \basecell{0.0359}{0.0011} & \basecell{0.0440}{0.0012} & \basecell{0.0438}{0.0031} & \basecell{0.0655}{0.0034} & \basecell{0.0244}{0.0013} & \basecell{0.0300}{0.0015}\\
CMDL      & \basecell{0.0535}{0.0027} & \basecell{0.0759}{0.0028} & \basecell{0.0276}{0.0013} & \basecell{0.0368}{0.0015} & \basecell{0.0338}{0.0023} & \basecell{0.0569}{0.0025} & \basecell{0.0214}{0.0011} & \basecell{0.0260}{0.0012}\\
ITCoHD-MRec & \basecell{0.0567}{0.0021} & \basecell{0.0838}{0.0025} & \basecell{0.0313}{0.0011} & \basecell{0.0382}{0.0013} & \basecell{0.0426}{0.0020} & \basecell{0.0636}{0.0023} & \basecell{0.0237}{0.0011} & \basecell{0.0292}{0.0011}\\
\textcolor{red}{COHESION} & \revisioncell{0.0660}{0.0023} & \revisioncell{0.0984}{0.0025} & \revisioncell{0.0360}{0.0013} & \revisioncell{0.0438}{0.0014} & \revisioncell{0.0445}{0.0027} & \revisioncell{0.0664}{0.0029} & \revisioncell{0.0247}{0.0016} & \revisioncell{0.0306}{0.0017}\\
\midrule
\rowcolor{TableGroup}\multicolumn{9}{l}{\textit{MAGNET variants}}\\[-1pt]
\textcolor{red}{MAGNET-SV} & \mainranktwo{\revisioncell{0.0710}{0.0018}} & \mainranktwo{\revisioncell{0.1038}{0.0021}} & \mainranktwo{\revisioncell{0.0404}{0.0011}} & \mainranktwo{\revisioncell{0.0490}{0.0011}} & \mainranktwo{\revisioncell{0.0465}{0.0021}} & \mainranktwo{\revisioncell{0.0700}{0.0021}} & \mainranktwo{\revisioncell{0.0268}{0.0009}} & \mainranktwo{\revisioncell{0.0330}{0.0010}}\\
\textcolor{red}{\textbf{MAGNET-DV}} & \mainrankone{\revisionbest{0.0741\psig{**}}{0.0015}} & \mainrankone{\revisionbest{0.1072\psig{**}}{0.0017}} & \mainrankone{\revisionbest{0.0405\psig{**}}{0.0009}} & \mainrankone{\revisionbest{0.0499\psig{**}}{0.0011}} & \mainrankone{\revisionbest{0.0482\psig{*}}{0.0017}} & \mainrankone{\revisionbest{0.0719\psig{**}}{0.0019}} & \mainrankone{\revisionbest{0.0269\psig{*}}{0.0011}} & \mainrankone{\revisionbest{0.0331\psig{*}}{0.0011}}\\
\bottomrule
\end{tabular}}
\end{table}

\paragraph{Comparison scope and statistical testing.}
\rev{We apply a two-sided one-sample $t$-test to the five seed-wise performance differences between MAGNET-DV and the underlined strongest observed protocol-compatible non-MAGNET baseline in each cell ($n=5$, $df=4$), with $^{*}p<0.05$ and $^{**}p<0.01$, without family-wise correction. These tests condition on the official fixed split and quantify optimization variability rather than split-to-split uncertainty.} The cell-wise markers in \Cref{tab:main_results,tab:microlens_results} report these comparisons.

\begin{table}[!t]
\centering
\revisiontable
\caption{\rev{MicroLens-100K results over five matched seeds on one fixed split (mean $\pm$ sample standard deviation). Ranking and significance conventions follow \Cref{tab:main_results}.}}
\label{tab:microlens_results}
\small
\setlength{\tabcolsep}{11pt}
\renewcommand{\arraystretch}{1.08}
\begin{tabular}{lcc}
\toprule
Method & R@20 & N@20\\
\midrule
LightGCN  & \presult{0.0584}{0.0028} & \presult{0.0372}{0.0014}\\
MMGCN     & \presult{0.0706}{0.0034} & \presult{0.0391}{0.0019}\\
GRCN      & \presult{0.1058}{0.0018} & \presult{0.0449}{0.0012}\\
BM3       & \presult{0.0981}{0.0031} & \presult{0.0406}{0.0016}\\
FREEDOM   & \presult{0.1024}{0.0025} & \presult{0.0436}{0.0010}\\
GUME      & \presult{0.1087}{0.0027} & \presult{0.0465}{0.0015}\\
MGCN      & \mainrankthree{\psecond{0.1103}{0.0029}} & \mainrankthree{\psecond{0.0471}{0.0014}}\\
\midrule
MAGNET-SV & \mainranktwo{\presult{0.1162}{0.0021}} & \mainranktwo{\presult{0.0501}{0.0011}}\\
\textbf{MAGNET-DV} & \mainrankone{\pbest{0.1194\psig{**}}{0.0019}} & \mainrankone{\pbest{0.0522\psig{**}}{0.0012}}\\
\bottomrule
\end{tabular}
\end{table}

\begin{itemize}[leftmargin=*,itemsep=1.5pt,topsep=1.5pt]
\item[--] \textbf{Overall performance.}
Across the five datasets, MAGNET-DV exceeds the strongest reproduced protocol-compatible non-MAGNET baseline in every reported main-table cell. \rev{These seed-wise difference tests support the cell-wise differences at $p<0.05$ or $p<0.01$, as marked in \Cref{tab:main_results,tab:microlens_results}.}

\item[--] \textbf{Cross-domain performance.}
MicroLens differs from the Amazon domains in content type, interaction distribution, and recommendation context. Under its compatible comparison set, MAGNET's short-video result provides evidence beyond product images and descriptions.

\item[--] \textbf{Comparison with strong contrastive multimodal methods.}
Among the compared approaches, contrastive-learning baselines are consistently strong, especially on Clothing and Electronics.
MAGNET retains its empirical advantage against strong contrastive-learning baselines.

\item[--] \textbf{Dual-view versus single-view.}
Dual-view provides the stronger overall default across the reported cells, whereas single-view can remain close or slightly stronger on an isolated metric. These results suggest that structural augmentation is useful when content-induced connectivity is reliable.
\end{itemize}

Cross-model differences in \Cref{tab:main_results,tab:microlens_results} do not identify individual component contributions; \Cref{sec:vi_comparison,sec:ablation,sec:template-design-analysis} provide matched controls for structure, components, and expert design. The MicroLens comparison uses the protocol-compatible methods listed in \Cref{sec:baselines}.

\subsection{\rev{Tail-Item and Low-History User Performance}}
\label{sec:bucketed}
We compute buckets from the training split only, preventing validation or test interactions from affecting membership. Items are sorted by training interaction count (ties are broken by item identifier), and the bottom 20\% of items form the tail set. Users are sorted analogously by training-history length to define the bottom-20\% cold-user set. \rev{Here, ``cold-user'' denotes a low-history warm-start subset defined from the training split, rather than a zero-interaction cold-start setting.} Tail-item Recall@20 is computed over test instances whose held-out target belongs to the tail set; cold-user NDCG@20 is computed over test instances belonging to cold users. We report the four non-MAGNET baselines reproduced across both the Amazon and MicroLens evaluations under the shared protocol, providing a consistent comparator set across all five datasets.

\Needspace{5\baselineskip}
\begingroup\interlinepenalty=10000
\rev{We evaluate the same five frozen checkpoints used in the main comparison, without bucket-specific retraining or selection, and report their bucket-wise means as descriptive summaries; formal significance claims remain confined to the full-test cell-wise comparisons in \Cref{tab:main_results,tab:microlens_results}.}\par
\endgroup

\begin{table}[!t]
\centering
\revisiontable
\caption{\rev{Descriptive means over the five frozen main-table checkpoints. Tail items and cold users are the bottom 20\% by training-split item frequency and user-history length, respectively; no significance claim is made. Best values are boldfaced.}}
\label{tab:bucketed_results}
\footnotesize
\setlength{\tabcolsep}{4.3pt}
\renewcommand{\arraystretch}{1.00}
\begin{tabular}{lccccc}
\toprule
\tablegrouprow
\multicolumn{6}{c}{\textbf{Tail-item Recall@20}}\\
\cmidrule(lr){1-6}
Method & Baby & Sports & Clothing & Electronics & MicroLens\\
\midrule
LightGCN & .0417 & .0495 & .0309 & .0307 & .0347\\
FREEDOM & .0629 & .0706 & .0616 & .0353 & .0695\\
MGCN & .0642 & .0733 & .0631 & .0406 & .0748\\
GUME & .0708 & .0781 & .0684 & .0434 & .0763\\
\midrule
\textbf{MAGNET-DV} & \textbf{.0760} & \textbf{.0827} & \textbf{.0738} & \textbf{.0473} & \textbf{.0831}\\
\addlinespace[2pt]
\cmidrule(lr){1-6}
\tablegrouprow
\multicolumn{6}{c}{\textbf{Cold-user NDCG@20}}\\
\cmidrule(lr){1-6}
Method & Baby & Sports & Clothing & Electronics & MicroLens\\
\midrule
LightGCN & .0274 & .0321 & .0196 & .0208 & .0305\\
FREEDOM & .0358 & .0404 & .0351 & .0221 & .0361\\
MGCN & .0363 & .0418 & .0359 & .0252 & .0392\\
GUME & .0393 & .0444 & .0393 & .0263 & .0401\\
\midrule
\textbf{MAGNET-DV} & \textbf{.0418} & \textbf{.0468} & \textbf{.0430} & \textbf{.0289} & \textbf{.0436}\\
\bottomrule
\end{tabular}
\vspace{2pt}

{\scriptsize\parbox{0.96\linewidth}{\centering Eligible test instances (tail target / cold user): Baby 1,842/3,214; Sports 3,506/5,792; Clothing 3,221/6,087; Electronics 18,436/27,905; MicroLens 9,618/16,742.}}
\end{table}

As shown in \Cref{tab:bucketed_results}, MAGNET achieves the highest descriptive mean within the reported comparison set for both tail-item retrieval and cold-user (low-history) ranking. Eligible test counts for every bucket are reported in the table footer.

\subsection{Analysis of Content-Induced Structural Augmentation}
\label{sec:vi_comparison}
Content-induced interactions can be introduced either as an overlay on one propagation graph or as a separately encoded structural view. All rows below share MAGNET's calibrated expert bank, router, recommendation and routing objectives, and checkpoint-selection protocol. Graph construction and view encoding vary, and cross-view alignment is used only in the dual-view configuration. We compare the observed graph alone, an unfiltered content overlay, a \rev{VI-MMRec-style filtered overlay}~\cite{Xu2026VIMMRec}, and MAGNET's jointly encoded dual view. The focused overlay diagnostic reports Baby and Sports; the matched MAGNET-SV/DV comparison over all five benchmarks appears in \Cref{tab:main_results,tab:microlens_results}.

For both overlay controls, candidates are collected from the union of the top-$k$ visual and textual neighbors of items in the training history, with $k=20$, excluding observed edges and removing duplicates. The \emph{unfiltered overlay} adds all candidate edges to a single propagation graph. The \emph{filtered overlay} ranks the same candidates by $c_{u,j}$ in \Cref{eq:augmentation-score} and retains at most $r=200$ candidates per user. The dual-view configuration uses the same filtered edge set, but encodes the observed and observed-plus-augmented graphs independently before fixed mean fusion.

\begin{table}[!t]
\centering
\revisiontable
\caption{Structural-augmentation controls under a common MAGNET scoring architecture and evaluation protocol. All rows use diagnostic seed~9; graph construction and view encoding vary, with cross-view alignment applied only to the dual-view row. The comparison is descriptive and is not used for significance testing.}
\label{tab:vi_plugin}
\footnotesize
\setlength{\tabcolsep}{5.0pt}
\renewcommand{\arraystretch}{1.06}
\resizebox{0.94\linewidth}{!}{%
\begin{tabular}{llcccc}
\toprule
\multirow{2}{*}{Graph construction} & \multirow{2}{*}{Encoding} & \multicolumn{2}{c}{Baby} & \multicolumn{2}{c}{Sports}\\
\cmidrule(lr){3-4}\cmidrule(lr){5-6}
& & R@20 & N@20 & R@20 & N@20\\
\midrule
Observed graph only & Single view & .1067 & .0476 & .1181 & .0549\\
Unfiltered overlay & Single graph & .1068 & .0475 & .1183 & .0551\\
Filtered overlay~\cite{Xu2026VIMMRec} & Single graph & .1072 & .0478 & .1185 & \textbf{.0553}\\
\midrule
\textbf{Joint observed/augmented} & \textbf{Dual view} & \textbf{.1090} & \textbf{.0483} & \textbf{.1186} & .0552\\
\bottomrule
\end{tabular}}
\end{table}

Under single-graph encoding, the filtered overlay is numerically higher than the unfiltered overlay in all four displayed cells. \rev{The dual-view variant is highest in three of the four cells, while the filtered single-graph variant is marginally higher on Sports NDCG@20 (0.0553 versus 0.0552).} Together with the five-seed MAGNET-SV/DV comparisons in \Cref{tab:main_results,tab:microlens_results}, these results support the view that controlled structural augmentation can complement routing in the evaluated settings.

\subsection{Component Ablation Study}
\label{sec:ablation}

To quantify the contribution of major components in MAGNET, we conduct an ablation study on four datasets (Baby, Sports, Clothing, and Electronics).

Specifically, we consider the following variants, covering (i) view construction and cross-view alignment, (ii) progressive-routing schedules and stage-specific regularizers, and (iii) expert design choices (templates and MoE fusion):

\begin{itemize}[leftmargin=*,itemsep=1.5pt,topsep=2pt]
\item \textbf{w/o Routing Regularizers:} sets $\lambda_r=0$, removing both stage-specific routing objectives while keeping the dual-view backbone and MoE fusion unchanged.
\item \rev{\textbf{w/o Scale Calibration:} feeds the three source representations directly to the templates without the normalization in \Cref{eq:cue-calibration}.}
\item \textbf{w/o View-contrastive:} disables view-level alignment between the UI-view and UIG-view representations, while keeping the dual-view backbone and MoE fusion intact.
\item \textbf{Fixed-step Switch:} keeps the two-stage regularization design but replaces entropy-triggered switching with a manually scheduled transition at $t=T/2$.
\item \rev{\textbf{Coverage stage only:} keeps the Stage~1 coverage-promoting routing objective throughout training and disables the transition to Stage~2. It therefore never introduces instance-level specialization.}
\item \rev{\textbf{Instance sharpening from initialization:} removes Stage~1, the coverage trigger, and the coverage guard, and applies instance-entropy sharpening from the beginning of training. This control tests whether specialization is stable without the initial coverage-promoting stage.}
\item \rev{\textbf{Free mixtures:} removes the shared Dom/Bal/Com template structure while retaining the same nine expert branches, router, Top-$K$ budget, backbone, and training protocol. Each expert independently learns a three-way source mixture. This is the same model configuration reported as free learnable mixtures in \Cref{tab:template_design}; this table reports seed~9, whereas \Cref{tab:template_design} reports the five-seed mean and sample standard deviation.}
\item \textbf{w/o MoE Fusion:} replaces the MoE module with a simple non-expert fusion head ($\{\mathbf{z}, \mathbf{h}^{A}, \mathbf{h}^{S}\}$), eliminating sparse routing and expert specialization.
\end{itemize}

\begin{table}[!htbp]
\centering
\caption{Ablation of major components (seed~9, same fixed split and protocol). Best values are boldfaced; ties boldfaced jointly.}
\label{tab:ablation}
\footnotesize
\setlength{\tabcolsep}{6.5pt}
\renewcommand{\arraystretch}{1.08}
\resizebox{0.92\textwidth}{!}{%
\begin{tabular}{lcc@{\hspace{7pt}}cc@{\hspace{7pt}}cc@{\hspace{7pt}}cc}
\toprule
\multirow{2}{*}{\textbf{Method}} &
\multicolumn{2}{c}{\textbf{Baby}} &
\multicolumn{2}{c}{\textbf{Sports}} &
\multicolumn{2}{c}{\textbf{Clothing}} &
\multicolumn{2}{c}{\textbf{Electronics}} \\
\cmidrule(lr){2-3}\cmidrule(lr){4-5}\cmidrule(lr){6-7}\cmidrule(lr){8-9}
& R@20 & N@20 & R@20 & N@20 & R@20 & N@20 & R@20 & N@20 \\
\midrule
\tablegrouprow
\multicolumn{9}{l}{\textbf{Routing schedule}}\\
w/o Routing Regularizers  & .1060 & .0466 & .1168 & .0539 & .1033 & .0478 & .0694 & .0317 \\
Fixed-step Switch    & .1078 & .0477 & .1180 & .0548 & .1060 & .0494 & .0714 & .0328 \\
\rev{Coverage stage only}  & \rev{.1071} & \rev{.0473} & \rev{.1177} & \rev{.0545} & \rev{.1056} & \rev{.0490} & \rev{.0710} & \rev{.0325} \\
\rev{Instance sharpening from initialization} & \rev{.1048} & \rev{.0458} & \rev{.1158} & \rev{.0529} & \rev{.1021} & \rev{.0470} & \rev{.0686} & \rev{.0311} \\
\addlinespace[1pt]
\tablegrouprow
\multicolumn{9}{l}{\textbf{Evidence preparation and auxiliary alignment}}\\
\rev{w/o Scale Calibration} & \rev{.1054} & \rev{.0463} & \rev{.1165} & \rev{.0537} & \rev{.1028} & \rev{.0475} & \rev{.0690} & \rev{.0314} \\
w/o View-contrastive & .1080 & .0479 & .1185 & \textbf{.0552} & .1071 & .0498 & .0720 & \textbf{.0331} \\
\addlinespace[1pt]
\tablegrouprow
\multicolumn{9}{l}{\textbf{Expert structure}}\\
\rev{Free mixtures} & \rev{.1061} & \rev{.0473} & \rev{.1174} & \rev{.0538} & \rev{.1040} & \rev{.0488} & \rev{.0702} & \rev{.0318} \\
w/o MoE Fusion       & .1039 & .0453 & .1150 & .0525 & .0999 & .0459 & .0675 & .0304 \\
\midrule
\tablefocusrow
\textbf{MAGNET}   & \textbf{.1090} & \textbf{.0483} & \textbf{.1186} & \textbf{.0552} & \textbf{.1078} & \textbf{.0502} & \textbf{.0721} & \textbf{.0331} \\
\bottomrule
\end{tabular}
}
\end{table}

\FloatBarrier
\Needspace{8\baselineskip}
Four grouped findings summarize the same-seed ablations relative to full MAGNET:
\begin{itemize}[leftmargin=*,itemsep=1.5pt,topsep=1.5pt]
\item[--] \textbf{Routing schedule.} All schedule controls remain below full MAGNET in the seed-9 diagnostic, with the fixed-step switch generally providing the closest alternative. \rev{Applying instance sharpening from initialization produces the largest degradation among the schedule controls.} The mechanism analyses in \Cref{sec:entropy_stage_analysis,sec:routing_sensitivity} further separate the roles of \rev{coverage-oriented regularization, instance uncertainty, and the entropy-triggered transition}.

\item[--] \rev{\textbf{Expert structure.} Replacing MoE with a single fusion head produces the largest overall ablation gap, while free mixtures without the shared template structure also remain below full MAGNET. Together with the matched comparison in \Cref{tab:template_design}, these controls support interaction-conditioned composition within a semantically organized expert space.}

\item[--] \rev{\textbf{Evidence preparation.} Removing scale calibration consistently lowers the displayed seed-9 results, consistent with calibration being important for preserving the intended triplet composition when heterogeneous source representations have different numerical scales.}

\item[--] \textbf{Auxiliary alignment.} Disabling view-contrastive learning produces the smallest overall ablation gap and ties full MAGNET on Sports NDCG@20 and Electronics NDCG@20. This pattern is consistent with view-level contrast serving as a lightweight auxiliary component rather than the primary performance driver.
\end{itemize}

\subsection{Structured Expert Design and Sensitivity}
\label{sec:hparam}
\label{sec:template-design-analysis}

\rev{Before varying individual template controls, we compare the structured bank with matched alternatives. The comparison asks whether semantic structure is useful and whether fixed coefficients provide a strong default relative to learnable alternatives.}

All controls retain the same $E{=}9$ expert branches, Top-$K$ budget, router architecture, and expert-head capacity. \emph{Homogeneous experts} retain the same nine branches and independently parameterized $2d\!\rightarrow\!d\!\rightarrow\!1$ scalar heads as the structured bank; the only change is that every branch receives the uniform source triplet $(1/3,1/3,1/3)$. \emph{Free learnable mixtures} optimize an independent three-way softmax for each branch without semantic anchors or a KL prior; \rev{\emph{anchored learnable templates} use the parameterization and KL anchor in \Cref{sec:magnet-moe}}; and \emph{fixed structured templates} use the canonical bank throughout training. \rev{The behavior-only router zero-masks the appearance and semantic query slots while retaining the same $6d\!\rightarrow\!E$ gate and structured bank.} Learnable coefficient variants add $3E$ scalar logits; all other branch, router, and expert-head capacities are matched.

\begin{table}[H]
\centering
\revisiontable
\caption{\rev{Expert and router controls matched in branch count, Top-$K$ budget, router architecture, and expert-head capacity on five datasets (NDCG@20; mean $\pm$ sample standard deviation over five matched seeds).}}
\label{tab:template_design}
\footnotesize
\setlength{\tabcolsep}{5pt}
\renewcommand{\arraystretch}{1.08}
\resizebox{\textwidth}{!}{%
\begin{tabular}{lccccc}
\toprule
Design & Baby & Sports & Clothing & Electronics & MicroLens\\
\midrule
\rowcolor{TableGroup}\multicolumn{6}{l}{\textbf{\textit{Expert/router structure}}}\\[-1pt]
Homogeneous experts & \presult{.0459}{.0015} & \presult{.0532}{.0015} & \presult{.0463}{.0016} & \presult{.0315}{.0013} & \presult{.0486}{.0019}\\
Free learnable mixtures & \presult{.0469}{.0012} & \presult{.0541}{.0016} & \presult{.0481}{.0013} & \presult{.0320}{.0015} & \presult{.0508}{.0015}\\
\begin{tabular}[c]{@{}l@{}}Structured experts + behavior-only\\router\end{tabular} & \presult{.0471}{.0012} & \presult{.0546}{.0010} & \presult{.0485}{.0012} & \presult{.0322}{.0010} & \presult{.0513}{.0014}\\
\midrule
\rowcolor{TableGroup}\multicolumn{6}{l}{\textbf{\textit{Template parameterization}}}\\[-1pt]
Anchored learnable templates & \presult{.0477}{.0009} & \presult{.0552}{.0012} & \presult{.0495}{.0013} & \presult{.0328}{.0012} & \pbest{.0525}{.0013}\\
Fixed structured templates & \pbest{.0481}{.0009} & \pbest{.0554}{.0012} & \pbest{.0499}{.0011} & \pbest{.0331}{.0011} & \presult{.0522}{.0012}\\
\bottomrule
\end{tabular}
}
\vspace{2pt}

{\parbox{\linewidth}{\scriptsize\raggedright\textit{Note.} ``Anchored learnable'' optimizes template coefficients with the KL anchor in \Cref{sec:magnet-moe}. ``Free learnable mixtures'' is the same model configuration as ``Free mixtures'' in \Cref{tab:ablation}: \Cref{tab:ablation} reports seed~9, whereas this table reports mean $\pm$ sample standard deviation over five matched seeds.}}
\end{table}

\rev{\Cref{tab:template_design} shows that fixed templates have higher reported means on the four Amazon domains, while the anchored variant holds a small numerical edge on MicroLens. Across domains, both structured variants achieve higher five-seed mean NDCG@20 than homogeneous and free mixtures on all five datasets, supporting a compact structured coverage basis in the evaluated settings. The behavior-only router also trails the all-source router, supporting the use of content evidence in interaction-conditioned routing.}

Following the expert-template instantiation in \Cref{sec:magnet-moe}, we treat $(\alpha,\beta,\delta)$ as \emph{triplet-template controls}. They are shared at the type level and fixed during training. Specifically, $\alpha$ adjusts anchor strength in dominant-type experts, $\beta$ adjusts anchor bias in balanced-type experts, and $\delta$ allocates mass between the two auxiliary sources in complementary-type experts. These meanings are defined in the canonical order $[\mathrm{anchor},\mathrm{aux}_1,\mathrm{aux}_2]$ and are therefore valid for all three anchor groups.

We examine one template control at a time in independently evaluated one-dimensional diagnostic sweeps, while keeping the other controls at their default settings, and report R@10, R@20, and N@20. To emphasize response shape rather than absolute differences across independently executed sweeps, each metric is normalized to the evaluated default point within its own sweep. Family composition and sparse activation are studied separately in \Cref{sec:study_E_K}.

\begin{figure}[!htbp]
  \centering
  \includegraphics[width=\linewidth]{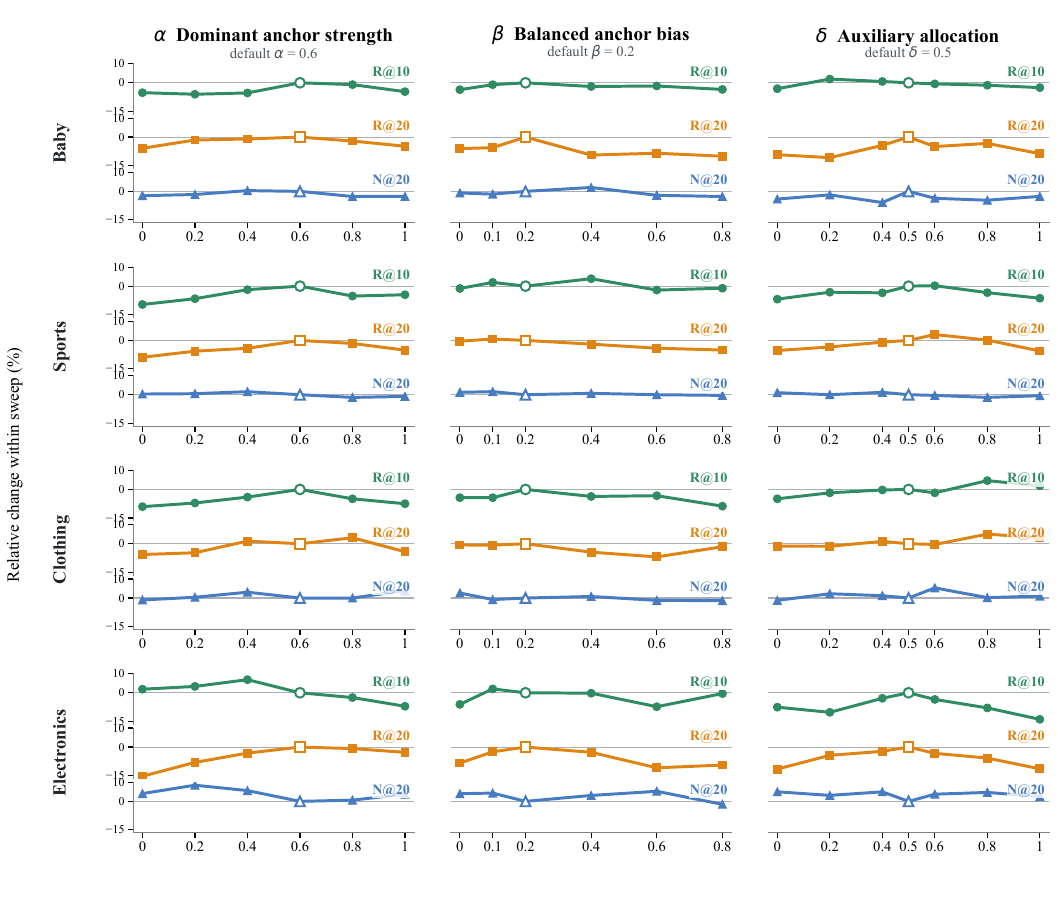}
  \Description{Twelve dataset-by-control panels, each with separate Recall at 10, Recall at 20, and NDCG at 20 response lanes, as the three triplet-template controls vary on Baby, Sports, Clothing, and Electronics.}
  \caption{Independent seed-9 diagnostic sweeps of MAGNET-DV over the triplet-template controls $(\alpha,\beta,\delta)$. Markers report evaluated grid points, with connecting segments included only as visual guides. Within each one-dimensional sweep, every metric is normalized to that sweep's evaluated default setting; hollow markers identify the corresponding default points $(0.6,0.2,0.5)$. R@10, R@20, and N@20 occupy separate lanes with a common relative-change scale.}
  \label{fig:para12}
\end{figure}

As shown in \Cref{fig:para12}, we make the following observations:
\begin{itemize}[leftmargin=*,itemsep=1.5pt,topsep=1.5pt]
  \item[--] \textbf{Stable shared operating region.} Around the evaluated defaults, the within-sweep responses generally change gradually across all four datasets. More extreme settings can incur larger declines---most visibly on Electronics---so the figure supports a broad common operating region rather than uniform insensitivity over the entire sweep.

  \item[--] \textbf{Distinct template roles.} The dominant strength $\alpha$ shows a mild mid-range preference, while the balanced bias $\beta$ remains stable over small-to-moderate values. The auxiliary allocation $\delta$ is more dataset dependent, consistent with domain-specific complementary evidence.

  \item[--] \textbf{Lightweight adaptation.} The evaluated defaults lie within broad stable regions; when moving beyond the Amazon domains, a validation-only one-dimensional sweep provides a lightweight check of the template controls.
\end{itemize}

\begin{figure*}[t]
  \centering
  \includegraphics[width=\linewidth]{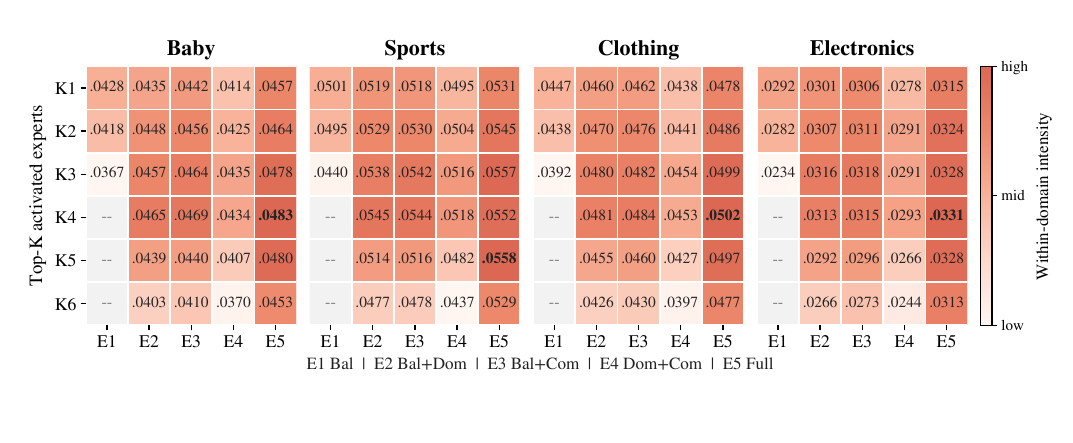}
  \Description{Four heatmaps showing NDCG at 20 across five expert-family compositions and six Top-K activation levels on the four Amazon datasets.}
  \caption{Fixed-seed (seed~9) family-composition and Top-$K$ diagnostic sweep (NDCG@20). E1--E5 denote Bal, Bal+Dom, Bal+Com, Dom+Com, and the full Dom+Bal+Com bank; K1--K6 denote the number of activated experts. Gray cells marked ``--'' are infeasible settings with $K>E$. Color intensity is normalized within each dataset and is not comparable across datasets; exact values are printed in every valid cell. E5/K4 matches the default architecture; all cells report fixed-seed NDCG@20 and are separate from the five-seed means in \Cref{tab:main_results}. Bold values mark the numerical maximum within each displayed sweep; ties are boldfaced jointly.}
  \label{fig:expert}
\end{figure*}

\subsection{Family Composition and Sparse Activation}
\label{sec:study_E_K}

We study expert-family composition and the number of activated experts $K$ (Top-$K$ routing).
\rev{This study complements the template controls by examining how expert-family composition and sparse-activation level jointly affect performance.}
(Here, Top-$K$ denotes the number of activated experts in MoE routing; the $N$ in R@$N$/N@$N$ is the evaluation cutoff.)

\noindent\textbf{Experimental setup.}
We visualize $N@20$ in \Cref{fig:expert} as a compact summary of the two-dimensional control.
We sweep $K\in\{1,2,3,4,5,6\}$ (shown as K1--K6), and invalid configurations with $K>E$ are marked as ``--''.

We vary the \emph{expert-family composition} with $E\le 9$.
Since each template family (Dom/Bal/Com) instantiates one expert per evidence-source group (three groups in total) following \Cref{sec:magnet-triplet-experts}, a single family yields $E{=}3$ experts, any two families yield $E{=}6$, and the full set yields $E{=}9$.
Accordingly, the x-axis labels E1--E5 in \Cref{fig:expert} correspond to five fixed compositions:
Bal-only ($E{=}3$), Bal+Dom ($E{=}6$), Bal+Com ($E{=}6$), Dom+Com ($E{=}6$), and Full (Bal+Dom+Com, $E{=}9$).

Unless specified, we keep the full MAGNET training protocol unchanged and vary only family composition and Top-$K$ activation.
Results are shown in \Cref{fig:expert}; we obtain the following findings:

\begin{itemize}[leftmargin=*,itemsep=1.5pt,topsep=1.5pt]
  \item[--] \textbf{The full bank remains near-best in the joint family-count/Top-$K$ diagnostic.}
  The Full setting (E5, $E{=}9$) remains near-best across datasets in the favorable $K$ region. \rev{\Cref{tab:template_design} provides the capacity-matched structured-versus-unstructured comparison.}

  \item[--] \textbf{Including Bal improves robustness under partial compositions.}
  Among the $E{=}6$ settings, combinations that include Bal (E2/E3) are generally more stable than Dom+Com (E4), suggesting that balancing-type experts help avoid brittle routing when capacity is limited.

  \item[--] \textbf{Top-$K$ is stable over a moderate-sparsity plateau.}
  For most configurations with $E\ge 6$, performance improves from small $K$ values and remains strong over a broad K3--K5 plateau. We therefore use $K{=}4$ as the shared default near the center of this stable region.
\end{itemize}

\begin{figure*}[t]
  \centering
  \includegraphics[width=0.91\textwidth]{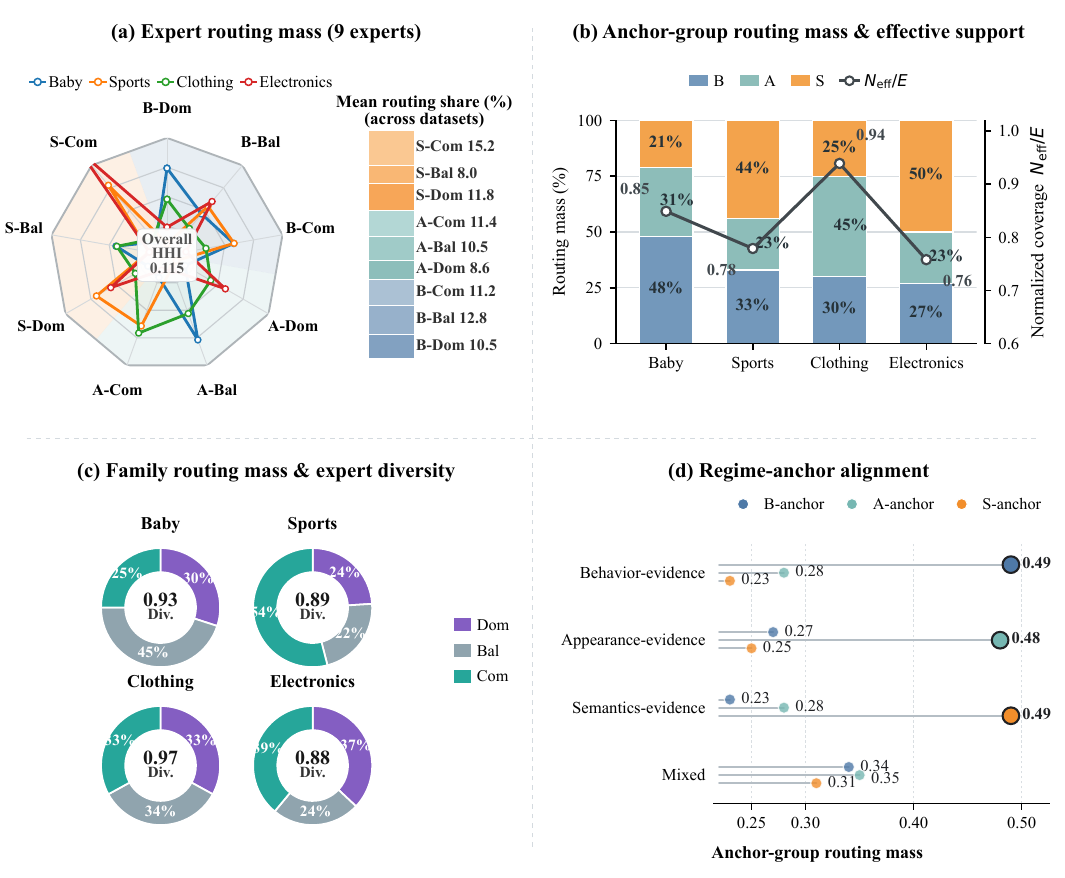}
  \Description{Four dense-routing diagnostics: expert routing-mass radar, anchor-group routing-mass shares with effective support, expert-family routing-mass mixture and diversity, and regime-conditioned anchor-group routing mass.}
  \caption{Fixed-seed (seed 9) routing diagnostics under the default 9-expert configuration.
    (a) Dense pre-Top-$K$ routing mass over the full expert pool, with background sectors and the right inset summarizing cross-dataset mean routing shares; (b) anchor-group routing-mass shares and effective support;
    (c) expert-family composition and diversity, where Div is computed from the normalized entropy of the full nine-expert mean routing distribution in \Cref{eq:routing-diagnostics}, while the three aggregated family shares are shown separately; and \rev{(d) Regime-conditioned anchor-group routing mass. For each Amazon domain and probe-defined regime, anchor-group mass is averaged over the corresponding diagnostic interactions; the figure reports the arithmetic mean of the four domain-level values. The legend follows the order B, A, S}. Anchor-group mass is the summed routing mass over experts sharing the same anchor label.}
  \label{fig:routing_combined}
\end{figure*}

\subsection{Routing Analysis of the Nine-Expert Pool}
\label{sec:routing_diagnostics}

We report routing statistics for the default $E{=}9$, $K{=}4$ configuration, using the seed-9 MAGNET-DV checkpoint under the same evaluation split as the main results. For each interaction, the router produces a dense distribution $\boldsymbol{\pi}_{ui}$ before Top-$K$ sparsification. The statistics below use this dense distribution and aggregate it over the evaluation split; the training-time switch in \Cref{sec:entropy} instead evaluates the dense mini-batch mean at each optimization step and uses a $W$-step persistence counter. Let $\bar{\boldsymbol{\pi}}$ denote the resulting dataset-level mean:
\begin{equation}
\label{eq:mean-routing}
\bar{\boldsymbol{\pi}}
=
\mathbb{E}_{(u,i)\sim \mathcal{D}_{\mathrm{eval}}}\!\left[\boldsymbol{\pi}_{ui}\right],
\qquad
\sum_{e=1}^{E}\bar{\pi}_{e}=1.
\end{equation}

We summarize coverage with the effective expert count $N_{\mathrm{eff}}$ and characterize concentration and uniformity with $\mathrm{HHI}$ and $\mathrm{Div}$:
\begin{equation}
\label{eq:routing-diagnostics}
\begin{aligned}
H(\bar{\boldsymbol{\pi}})&=-\sum_{e=1}^{E}\bar{\pi}_{e}\log \bar{\pi}_{e},
&N_{\mathrm{eff}}&=\exp\!\big(H(\bar{\boldsymbol{\pi}})\big),\\
\mathrm{HHI}(\bar{\boldsymbol{\pi}})&=\sum_{e=1}^{E}\bar{\pi}_{e}^{2},
&\mathrm{Div}(\bar{\boldsymbol{\pi}})&=\frac{H(\bar{\boldsymbol{\pi}})}{\log E}\in[0,1].
\end{aligned}
\end{equation}
Here, $\mathrm{HHI}$ increases with concentration, whereas $\mathrm{Div}$ increases with uniformity. Anchor-group and family shares summarize routing allocation over their corresponding labeled expert groups.

\begin{itemize}[leftmargin=*,itemsep=1.5pt,topsep=1.5pt]
  \item[--] \textbf{Broad routing-mass coverage.} \Cref{fig:routing_combined}(a)--(c) show low concentration and broad effective support in the dense routing distribution, with no single expert or family monopolizing the routed mass.

  \item[--] \textbf{Structured domain variation.} Anchor-group and family routing patterns vary coherently across datasets: Baby is more behavior-anchored and balanced, Sports uses complementary fusion more heavily, Clothing has the broadest expert diversity, and Electronics is more semantics-leaning and selective. These differences illustrate domain-dependent operation within one shared expert bank.

  \item[--] \rev{\textbf{Regime--anchor label alignment.} The regime-conditioned analysis below connects these dataset-level patterns to individual interactions by comparing probe-defined evidence regimes with routed anchor mass.}
\end{itemize}

\paragraph{Regime-conditioned routing.}
We reuse the independent assignments from \Cref{sec:diagnostic-protocol} as post-hoc strata and summarize MAGNET's anchor-group routing mass over the diagnostic interactions from the four Amazon domains within each stratum. \rev{As shown in \Cref{fig:routing_combined}(d), interactions in the behavior-, appearance-, and semantics-evidence regimes place the largest mass on the correspondingly anchored group, whereas Mixed interactions retain a flatter allocation. This agreement describes alignment between probe-defined regimes and expert-group anchor labels.}

\rev{The alignment is retained throughout the preset robustness range $\kappa\in\{0.40,0.45,0.50\}$: each source-specific evidence stratum preserves its matching anchor as the largest group, with a maximum pairwise cell deviation of 0.024.}

\subsection{Entropy-Triggered Routing Stage Analysis}
\label{sec:entropy_stage_analysis}

We analyze the two entropy statistics using a fixed-seed (seed 9) diagnostic trajectory on Electronics, the largest Amazon domain, together with the dense routing distribution before Top-$K$ truncation. \rev{Stage~2 begins only when the population-level coverage condition $H_{\mathrm{cov}}\ge H^*$ persists for $W$ steps.}

\paragraph{Mechanism dynamics on a representative dataset.}

For a batch of dense routing distributions, we define three normalized diagnostics:
\begin{equation}
\begin{aligned}
C_{\rm conc}&=\frac{1}{|\mathcal B|}\sum_{(u,i)\in\mathcal B}
\frac{\sum_e \pi_{ui,e}^{2}-1/E}{1-1/E},\\
M_3&=\frac{1}{|\mathcal B|}\sum_{(u,i)\in\mathcal B}\sum_{e\in\operatorname{Top3}(\boldsymbol\pi_{ui})}\pi_{ui,e},\qquad
L_{\max}=\max_e\frac{1}{|\mathcal B|}\sum_{(u,i)\in\mathcal B}\pi_{ui,e}.
\end{aligned}
\label{eq:routing-mechanism-diagnostics}
\end{equation}
Here $C_{\rm conc}$ ranges from zero for uniform assignments to one for one-hot assignments, $M_3$ is the Top-3 routing mass, and $L_{\max}$ is the maximum population routing mass assigned to any expert. All three summarize the dense pre-Top-$K$ routing probabilities; stage switching remains governed by $H_{\rm cov}$ in \Cref{eq:stage_switch}.

\begin{figure}[!t]
    \centering
    \includegraphics[width=\textwidth]{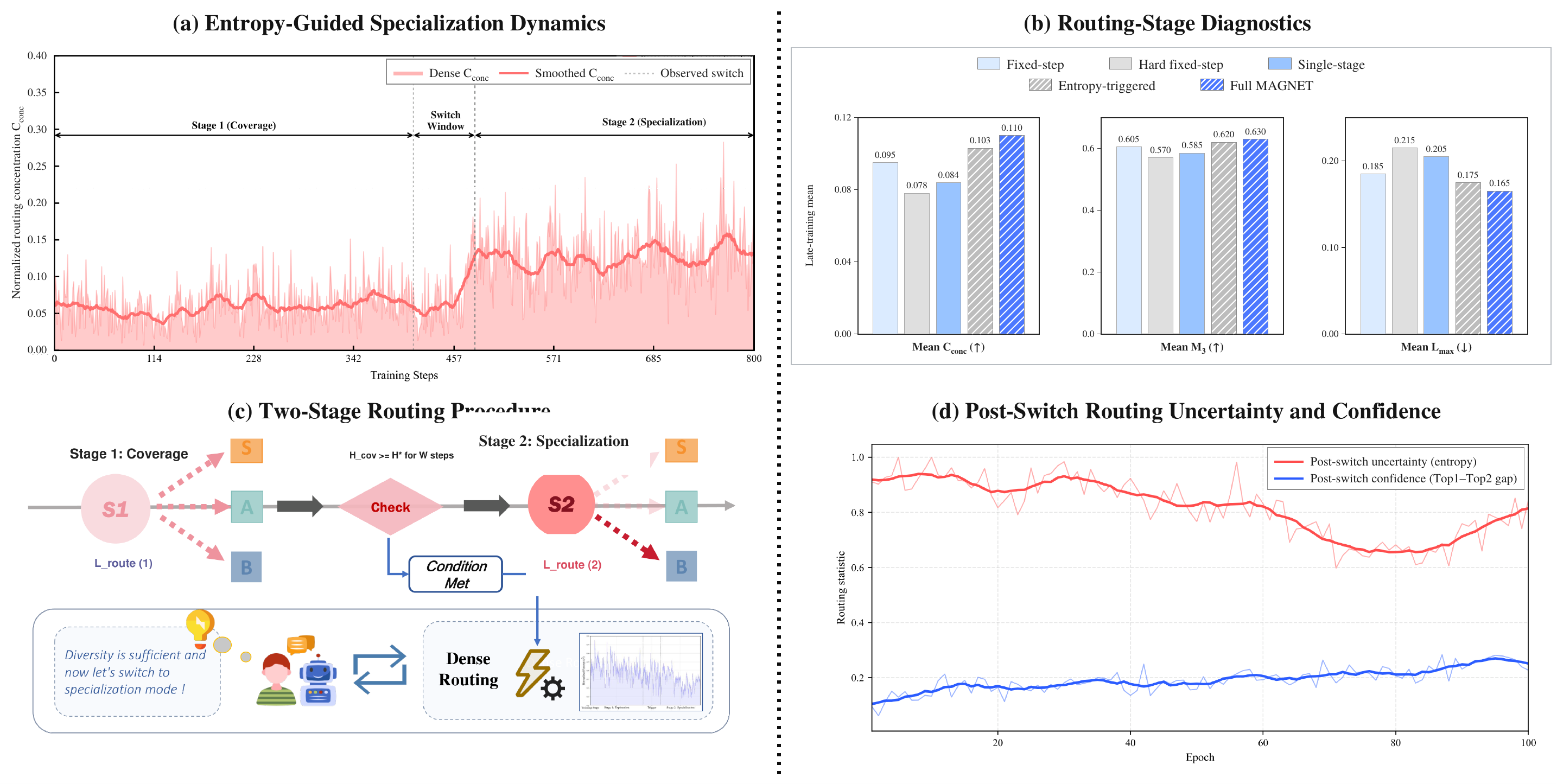}
    \Description{Four panels illustrating the entropy-triggered routing mechanism: concentration around the switch, routing-control comparisons, the two-stage procedure, and post-switch uncertainty and confidence trajectories.}
    \caption{\textbf{Fixed-seed (seed 9) entropy-triggered routing mechanism on Electronics.} (a) Normalized routing concentration in an update-level window around the coverage-triggered switch. (b) Late-training means of concentration, Top-3 mass, and maximum expert load over the final 10\% of optimization updates \rev{under the five schedule controls defined in the text}. (c) Procedural view of the two-stage transition with stage-specific routing objectives and the persistent coverage trigger. \rev{(d) Post-switch mean instance entropy and mean Top-1/Top-2 dense-routing margin over training epochs.}}
    \label{fig:routing_mechanism}
\end{figure}

\Cref{fig:routing_mechanism} provides a descriptive view of the transition on Electronics. Panel (a) tracks $C_{\rm conc}$ around the coverage-triggered switch. \rev{Panel (b) compares $(C_{\rm conc},M_3,L_{\max})$ under five schedule controls: a fixed-step switch that replaces the coverage trigger by $t=T/2$ while retaining the complete stage objectives and transient Stage~1 exploration weighting; a hard fixed-step two-stage schedule that also switches at $t=T/2$ but omits the transient Stage~1 exploration weighting; a single-stage entropy-regularized control that minimizes $\lambda_{\rm cov}(1-H_{\rm cov})+\lambda_{\rm MI}H_{\rm inst}$ throughout training without an explicit stage switch; an entropy-triggered two-stage schedule with fixed stage-specific objectives but no transient Stage~1 exploration weighting; and full MAGNET with the complete coverage trigger, transient Stage~1 exploration weighting, and coverage guard.}

Panel (c) summarizes the two-stage objective, and \rev{panel (d) shows mean instance uncertainty falling as the Top-1/Top-2 margin rises}. The trajectories show broad routing-mass coverage in Stage~1 and sharper routing with maintained population-level coverage in Stage~2; \Cref{tab:entropy_weight_strategies} reports the corresponding entropy-statistic controls.

\subsection{Controls and Sensitivity of Entropy-Triggered Routing}
\label{sec:routing_sensitivity}

Beyond the training trajectories, we test whether separating the two entropy statistics is empirically beneficial and whether the shared trigger and routing strength occupy a stable operating region.

\subsubsection{Coverage versus Instance Entropy}
\label{sec:coverage_vs_instance}

We compare four Stage~2 formulations while keeping Stage~1, the coverage-based trigger, model capacity, and validation protocol unchanged. Unlike the stage-removal controls in \Cref{tab:ablation}, all controls in \Cref{tab:entropy_weight_strategies} retain the two-stage schedule and the one-way transition triggered by $H_{\rm cov}$. They differ only in the statistic and constraint used after the transition.

\emph{Population-entropy proxy} uses the population-level entropy $H_{\rm cov}$ as a surrogate for local routing confidence, minimizing $\lambda_{\rm MI}H_{\rm cov}$ after the switch. \emph{Two-stage, instance entropy only} instead minimizes the appropriate per-instance uncertainty $\lambda_{\rm MI}H_{\rm inst}$, but provides no population-level diversity objective. \rev{\emph{Two entropies, no coverage guard} optimizes $\lambda_{\rm MI}(H_{\rm inst}-H_{\rm cov})$, jointly encouraging local decisiveness and global diversity. The full separated objective additionally retains the explicit coverage-preservation term $\lambda_{\rm guard}(1-H_{\rm cov})$.} The same coefficient magnitude $\lambda_{\rm MI}$ is reused in the first three Stage~2 variants.

\begin{table}[H]
\centering
\revisiontable
\caption{Stage~2 routing-statistic controls (seed~9; R@20/N@20). Stage~1, coverage trigger, capacity, and validation protocol are shared across rows. Best values are boldfaced.}
\label{tab:entropy_weight_strategies}
\footnotesize
\setlength{\tabcolsep}{4pt}
\resizebox{\linewidth}{!}{\begin{tabular}{lcccccccc}
\toprule
\multirow{2}{*}{Routing statistic/objective} & \multicolumn{2}{c}{Baby} & \multicolumn{2}{c}{Sports} & \multicolumn{2}{c}{Clothing} & \multicolumn{2}{c}{Electronics}\\
\cmidrule(lr){2-3}\cmidrule(lr){4-5}\cmidrule(lr){6-7}\cmidrule(lr){8-9}
& R@20 & N@20 & R@20 & N@20 & R@20 & N@20 & R@20 & N@20\\
\midrule
Population entropy as confidence proxy & .1054 & .0463 & .1160 & .0535 & .1030 & .0475 & .0690 & .0312\\
Two-stage, instance entropy only & .1073 & .0475 & .1179 & .0548 & .1058 & .0491 & .0713 & .0326\\
Two entropies, no coverage guard & .1081 & .0479 & .1184 & .0551 & .1070 & .0498 & .0720 & \textbf{.0332}\\
\midrule
\textbf{Separated objective} & \textbf{.1090} & \textbf{.0483} & \textbf{.1186} & \textbf{.0552} & \textbf{.1078} & \textbf{.0502} & \textbf{.0721} & .0331\\
\bottomrule
\end{tabular}}
\end{table}

The controls form a progressive same-seed mechanism check. Treating population entropy as a local confidence proxy produces the lowest performance across all eight displayed cells. Replacing it with the appropriate per-instance entropy recovers a substantial portion of the gap. \rev{Introducing the two-entropy objective further improves all eight cells over the instance-entropy-only control by jointly encouraging local decisiveness and population-level diversity. Adding the explicit coverage guard yields a small additional numerical gain in seven of the eight cells, with a marginal reversal on Electronics NDCG@20 (0.0331 versus 0.0332).}

The progression indicates that separating population coverage from instance uncertainty contributes more strongly than the final coverage-guard adjustment, whose effect is small and not uniformly positive across the displayed metrics. With expert capacity, the coverage-based trigger, optimization seed, and recommendation objective held constant, the rows compare Stage~2 routing formulations directly. The entropy trigger is a data-dependent training control that replaces a manually specified transition step with a coverage-based criterion, \rev{yielding modest improvements over the fixed-step alternative across all four Amazon datasets in the seed-9 comparison of \Cref{tab:ablation}.}

\subsubsection{Sensitivity to $(H^{\ast}, W)$}
\label{sec:sensitivity_hw}

\begin{figure}[!t]
  \centering
  \includegraphics[width=0.90\textwidth]{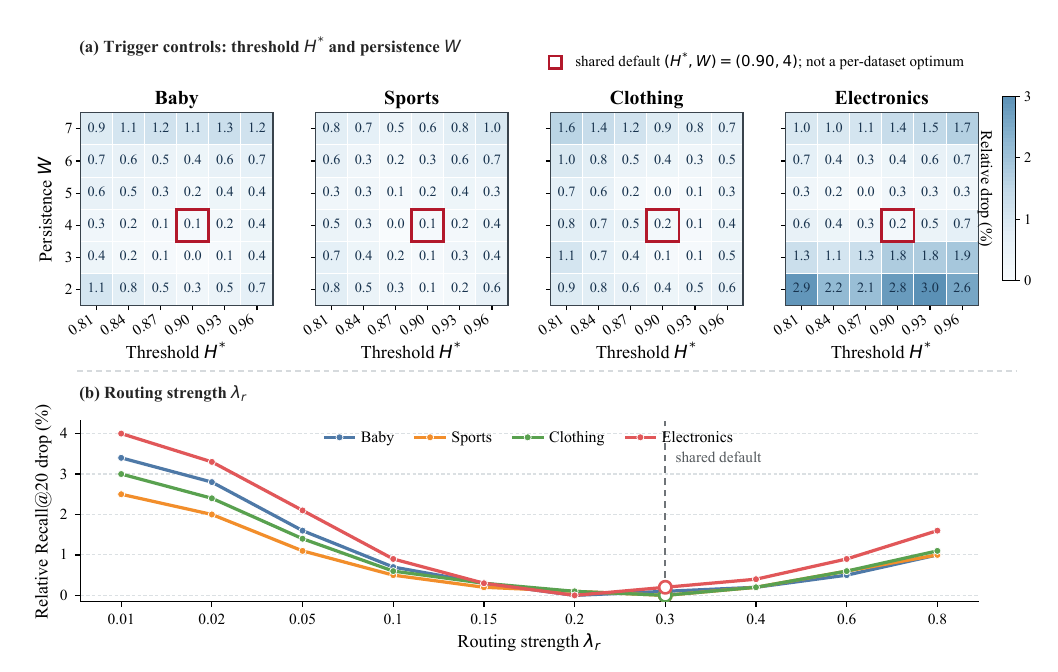}
  \Description{Four heatmaps of relative Recall at 20 drop over entropy threshold and persistence, followed by a shared line plot of routing-strength sensitivity.}
  \caption{Independent seed-9 diagnostic sweeps of the entropy-triggered schedule on four Amazon datasets. The top row reports the relative Recall@20 drop over $(H^*,W)$ for each dataset, and the shared lower panel reports the corresponding relative drop over $\lambda_r$. Each panel is normalized only within its own sweep, with relative drop computed from the best evaluated point for that dataset. The repeated red outline marks the shared preset $(0.90,4)$---not four per-dataset optima---and $(0.90,4,0.30)$ lies within the displayed stable regions.}
  \label{fig:routing_sensitivity}
\end{figure}

The heatmaps in \Cref{fig:routing_sensitivity} show mild degradation at extreme entropy thresholds or persistence windows. Because the same globally fixed configuration is evaluated on the four Amazon datasets in this sweep, the repeated red outline marks the shared preset rather than per-dataset optima; under within-domain normalization, it lies in a low-degradation region for all four.

\iffalse
\begin{itemize}
    \item[--] \textbf{Parameter meaning.} $H^{\ast}$ specifies the entropy level required to activate the stage transition, while $W$ controls how many consecutive steps must satisfy the trigger condition, reducing spurious switches caused by short-term fluctuations. Together, they regulate the switching timing and stability of the two-stage training process.

    \item[--] \textbf{Broad plateau indicates low sensitivity.} Across all four datasets, the heatmaps exhibit a stable high-performance region around moderate values of $(H^{\ast}, W)$, where Recall@20 varies only marginally. This suggests that the proposed schedule does not rely on delicate tuning and can generalize well across domains.

    \item[--] \textbf{Extreme settings lead to consistent degradation.} Overly small $W$ tends to make the trigger overly responsive, resulting in less stable switching behavior, while overly large $W$ or extreme $H^{\ast}$ values can delay the transition and weaken the intended stage-wise effect. Both cases produce a mild yet consistent performance drop at the search grid  boundaries.

    \item[--] \textbf{Dataset-specific optima with a shared robust region.} The best-performing $(H^{\ast}, W)$ may differ slightly across datasets, but the near-optimal regions largely overlap. Notably, the default configuration $(H^{\ast}, W)=(0.90,4)$ (used throughout the main tables) lies within this shared plateau, providing a reliable and fair choice without per-dataset tuning.
\end{itemize}
\fi

\subsubsection{Sensitivity to Routing Regularization Strength $\lambda_r$}
\label{sec:sensitivity_lambda_r}

We evaluate a wider fixed-seed $\lambda_r$ range than the main-configuration search set. We vary $\lambda_r$ over $\{0.01,\allowbreak 0.02,\allowbreak 0.05,\allowbreak 0.10,\allowbreak 0.15,\allowbreak 0.20,\allowbreak 0.30,\allowbreak 0.40,\allowbreak 0.60,\allowbreak 0.80\}$ while holding the shared trigger and routing ratios fixed. The response curves in \Cref{fig:routing_sensitivity} show under-regularization below $0.10$, a broad plateau around $0.15$--$0.40$, and mild regression under excessive regularization. \rev{The shared setting $\lambda_r=0.30$ lies within this stable region on all four Amazon datasets in this sweep.}

\iffalse
\begin{itemize}
    \item[--] \textbf{Under-regularization degrades performance.}
    When $\lambda_r \in [0.01, 0.05]$,  the model consistently underperforms across all datasets.
    This indicates that insufficient routing regularization weakens the intended stage-wise constraints for stable routing-mass coverage and specialization.

    \item[--] \textbf{A broad effective region yields robust performance.}
    As $\lambda_r$ increases, Recall@20 improves rapidly and reaches a stable high-performance plateau in the moderate range (roughly $0.15$--$0.40$).
    Within this region, performance differences are marginal, suggesting that the proposed routing scheme does not rely on delicate tuning of $\lambda_r$.

    \item[--] \textbf{Over-regularization causes mild regression.}
    Further increasing $\lambda_r$ to large values ($0.60$--$0.80$) leads to a small but consistent drop.
    This behavior aligns with the intuition that overly strong regularization may constrain router adaptivity and introduce optimization trade-offs with the main recommendation objective.
\end{itemize}

\begin{figure}[t]
    \centering
    \includegraphics[width=\linewidth]{figures/lamada_r.pdf}
    \caption{Sensitivity of routing regularization strength $\lambda_r$ on four datasets (Recall@20).}
    \label{fig:lambda_r}
\end{figure}
\fi

Together, the sweeps place the shared $(H^*,W,\lambda_r)$ setting within stable regions for both transition timing and routing strength.

\subsection{Representation-Space Diagnostics}
\label{sec:tsne}

We next use a qualitative representation-space visualization to inspect local
organization accompanying the aggregate routing patterns. Under matched plotting
conventions, the panels show category neighborhoods in the pre-fusion and
MAGNET-derived representations and source/family annotations of fused items.
For each observed interaction, the dense router weights and expert templates
yield effective B/A/S weights for fusing the calibrated source cues. We average
the resulting representations over interactions involving each item and over
each user's training history; dominant source and expert-family labels follow
the largest corresponding mean routing masses.

\begin{figure}[!t]
    \centering
    \includegraphics[width=0.88\textwidth]{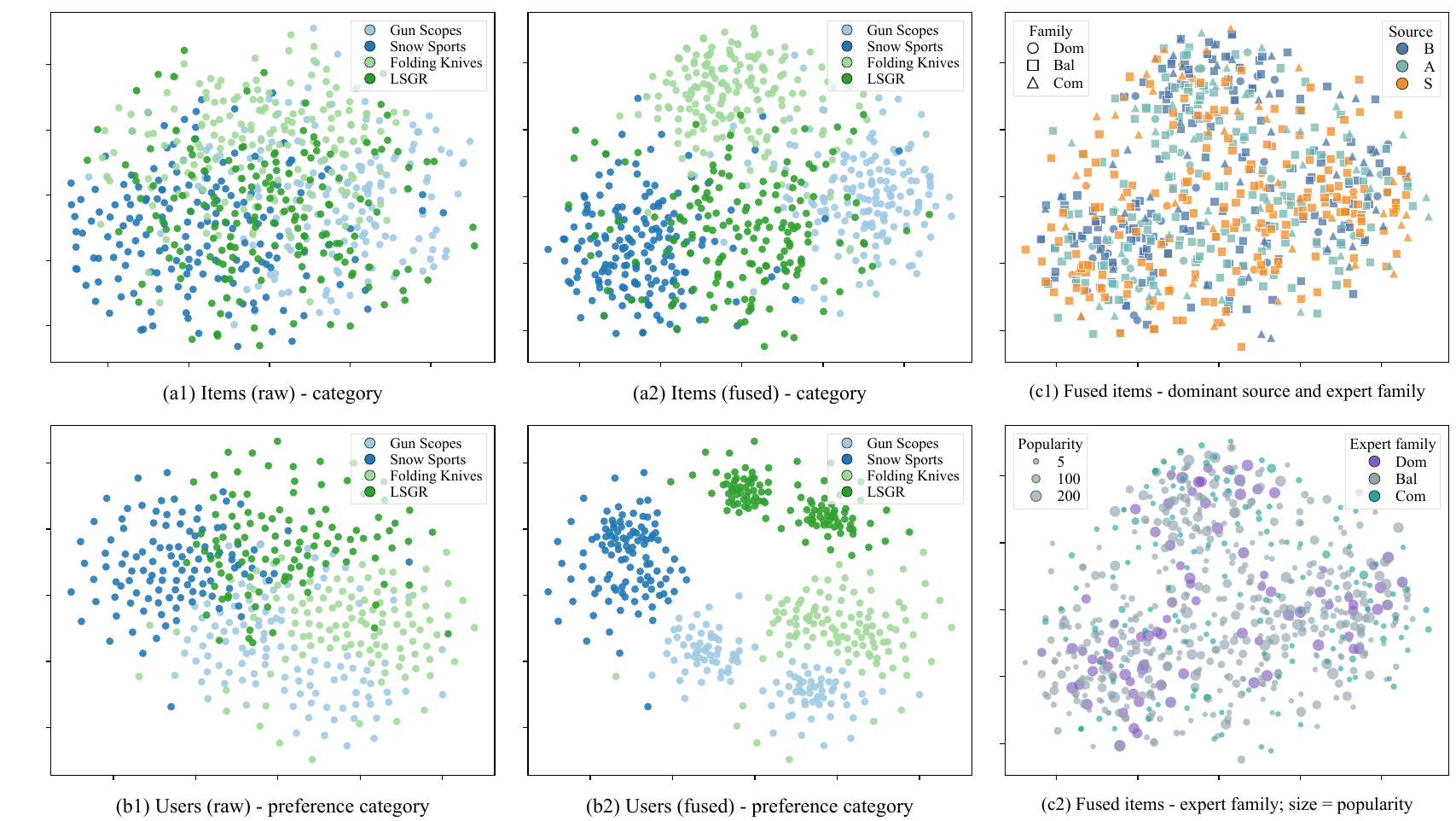}
    \Description{Six t-SNE scatter plots comparing pre-fusion and MAGNET-derived item and user representations, with fused items annotated by the dominant source and expert family obtained from mean routing masses over their observed training interactions; marker size additionally encodes item popularity.}
    \caption{Qualitative t-SNE visualization on the Sports dataset using the seed-9 MAGNET-DV checkpoint under matched plotting conventions. ``Raw'' and ``fused'' denote the pre-fusion and MAGNET-derived representations, respectively. Item-level source and family annotations aggregate routing masses over observed training interactions involving each item. LSGR denotes the Leisure Sports/Game Room category. Panels (c1)--(c2) show fused items annotated by dominant evidence source and expert family, with marker size in (c2) encoding training-set item popularity. Coordinate ticks are omitted because t-SNE axes have no cross-panel metric meaning.}
    \label{fig:tsne}
\end{figure}

\Cref{fig:tsne} visualizes local category neighborhoods and the descriptive
association of expert annotations with dominant evidence sources, expert
families, and popularity. Because t-SNE is nonlinear, we do not interpret
apparent cross-panel separation as evidence that one representation is
superior. Within this qualitative scope, the panels retain broad category
organization and show source/family annotations distributed across multiple
regions. This complements the aggregate routing masses in
\Cref{fig:routing_combined} without turning geometric proximity into a
performance claim.

\subsection{Efficiency}
\label{subsec:complexity_efficiency}

Message passing costs $O(L|\mathcal E|d)$ per view. The dense gate evaluates
all $E$ routing logits, while the expert-head computation scales with the
activated expert count $K$. Under identical hardware,
the dual-view model has the same trainable parameter count as MAGNET-SV because the two graph views share all embeddings and trainable modules; the additional view is a non-parametric sparse propagation path. It increases peak memory and time per epoch by 13.8\% and 10.2\% on Baby, and by 10.5\% and
6.4\% on Electronics, respectively (\Cref{tab:efficiency_summary}). MAGNET-SV
therefore remains the resource-oriented fallback. The realized sparse graph is
determined after candidate filtering and edge deduplication within the per-user
budget $r$.

The additional structure is concentrated in lightweight expert transformations and sparse routing: all nine branches share the evidence backbone, and only $K$ branches are active for an interaction. This accounting provides the basis for the accuracy--complexity trade-off.

The selected-epoch column contextualizes per-epoch cost: DV selects an earlier
checkpoint on Electronics, partly offsetting its additional propagation time.

\begin{table}[H]
\centering
\caption{Efficiency under the common hardware protocol. Time per epoch is
averaged over the training trajectory after a short warm-up; peak GPU memory
is the maximum allocated GPU memory observed during training. \rev{Relative time is normalized to SOIL within each
dataset. Params denotes the number of trainable scalar parameters (elements of tensors with
\texttt{requires\_grad=True}), reported in millions ($10^6$); frozen
feature tensors and non-trainable sparse graphs are excluded.}}
\label{tab:efficiency_summary}
\footnotesize
\setlength{\tabcolsep}{5pt}
\renewcommand{\arraystretch}{1.10}
\resizebox{\linewidth}{!}{%
\begin{tabular}{llrrrrr}
\toprule
\multirow{2}{*}{Dataset} & \multirow{2}{*}{Method} & \multirow{2}{*}{\rev{Params (M)$\downarrow$}} & \multicolumn{3}{c}{Per-epoch cost} & \multirow{2}{*}{Selected epoch} \\
\cmidrule(lr){4-6}
 & & & Peak memory (GB)$\downarrow$ & Time/epoch (s)$\downarrow$ & \rev{Rel. time} & \\
\midrule
\multirow{3}{*}{\textbf{Baby}}
 & SOIL      & \rev{2.0} & 2.44 & 1.49 & \rev{$1.00\times$} & 48 \\
\addlinespace[2pt]
 & MAGNET-SV & \rev{2.1} & \rev{3.11} & \rev{1.67} & \rev{$1.12\times$} & \rev{41} \\
 & MAGNET-DV & \rev{2.1} & \rev{3.54} & \rev{1.84} & \rev{$1.23\times$} & \rev{39} \\
\addlinespace[3pt]
\cmidrule(lr){1-7}
\multirow{3}{*}{\textbf{Electronics}}
 & SOIL      & \rev{16.6} & 6.18 & 20.60 & \rev{$1.00\times$} & 82 \\
\addlinespace[2pt]
 & MAGNET-SV & \rev{16.7} & \rev{7.06} & \rev{22.02} & \rev{$1.07\times$} & \rev{80} \\
 & MAGNET-DV & \rev{16.7} & \rev{7.80} & \rev{23.44} & \rev{$1.14\times$} & \rev{69} \\
\bottomrule
\end{tabular}}
\end{table}

\subsection{Case Study}
\label{sec:case_study}

\begin{figure}[!t]
  \centering
  \includegraphics[width=0.86\linewidth]{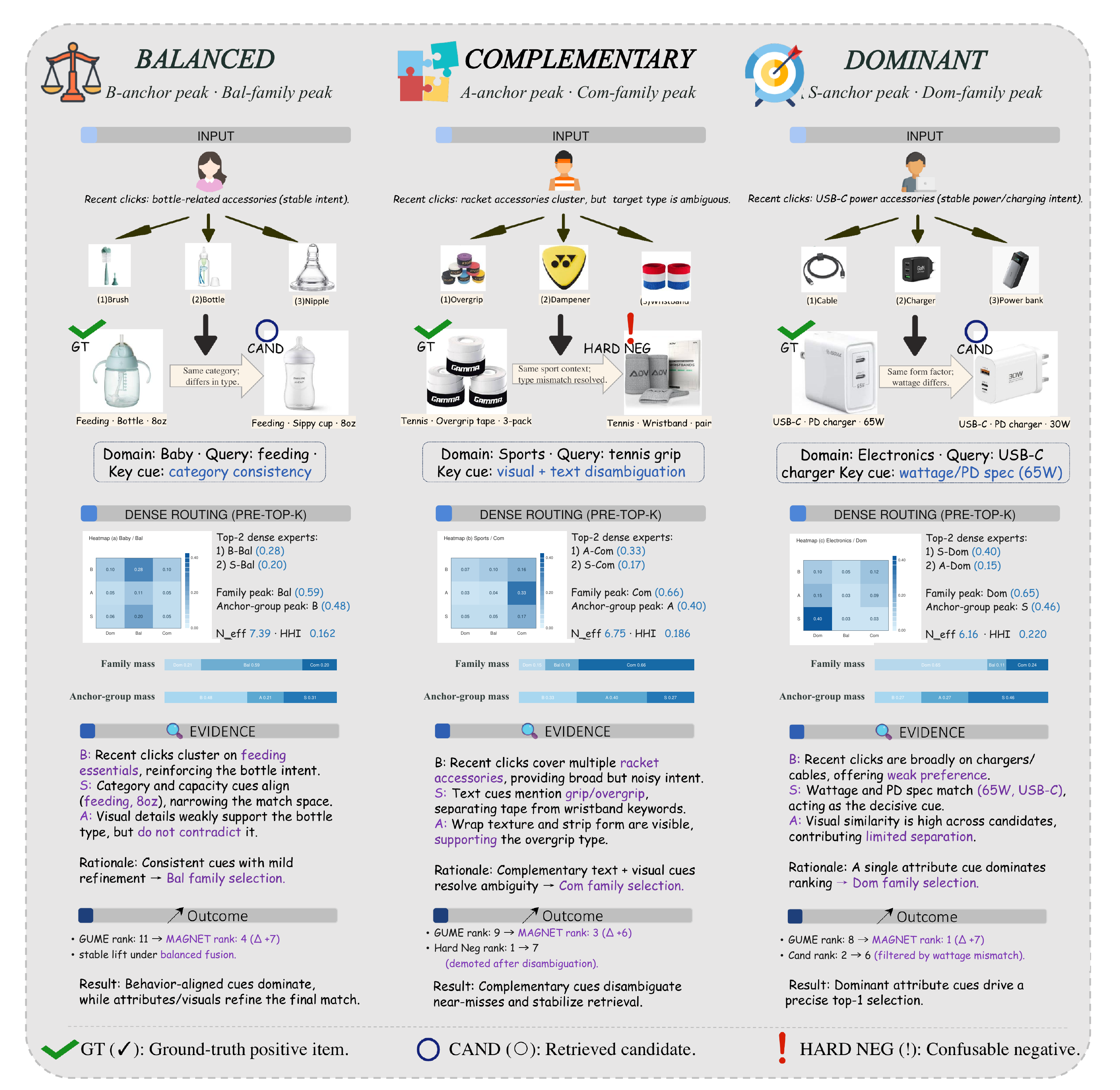}
  \Description{Three side-by-side qualitative recommendation cases illustrating balanced, complementary, and dominant routing. Each case shows recent interactions, a target and contrast item, the dense pre-Top-$K$ routing distribution, source evidence, and the resulting ranking change.}
  \caption{\textbf{Qualitative case study} spanning balanced, complementary, and dominant routing patterns. Each column preserves the complete case narrative---recent interactions and target contrast, normalized routing mass, source-level evidence, and ranking outcome---while the displayed routing summaries are computed from the same $3\times3$ mass grid. GUME is used as the comparison baseline in all three examples. \rev{Heatmaps show the dense pre-Top-$K$ routing distribution; forward scoring subsequently applies Top-$K$ selection and renormalization as defined in \Cref{eq:sparse-routing}.}}
  \label{fig:case_study}
\end{figure}

\paragraph{Construction and scope.}
The three illustrative examples span Bal, Com, and Dom routing patterns and illustrate evidence--routing--outcome narratives complementary to the aggregate analysis in \Cref{fig:routing_combined}. For each case, the displayed heatmap is normalized to unit routing mass, and all accompanying routing summaries are derived from that same $3\times3$ grid. \rev{The heatmap visualizes the dense pre-Top-$K$ routing distribution; forward scoring subsequently applies Top-$K$ selection and renormalization according to \Cref{eq:sparse-routing}.}

\paragraph{Key takeaways.}
We read \Cref{fig:case_study} as a qualitative illustration of the routing logic:
\begin{itemize}[leftmargin=*, itemsep=1.5pt, topsep=2pt]
  \item[--] \textbf{Balanced routing illustrates multi-source refinement.}
  In the illustrated \textbf{BALANCED} pattern, routing mass spreads across complementary sources: behavior anchors the user intent, while attributes and visual cues provide mild refinement without over-committing to a single cue.

  \item[--] \textbf{Complementary routing illustrates cross-source disambiguation.}
  In the illustrated \textbf{COMPLEMENTARY} pattern, no single source is sufficient; cross-source cues jointly distinguish a target from a near-miss.

  \item[--] \textbf{Dominant routing illustrates source-focused discrimination.}
  In the illustrated \textbf{DOMINANT} pattern, a decisive attribute-level constraint drives a peaked routing distribution while other signals play a supporting role.
\end{itemize}

The family names describe the prevailing routing pattern rather than mutually
exclusive decision rules: an interaction may retain nonzero mass in all three
families.

\section{Conclusion}
The regime-dependent crossovers identified in \Cref{sec:diagnostic-protocol} motivate interaction-conditioned fusion. MAGNET addresses this heterogeneity through routing over a calibrated, semantically structured fusion space. Dominant, balanced, and complementary experts make the covered fusion patterns explicit, while \rev{population-level coverage gates the transition to per-instance specialization and remains explicitly preserved thereafter}. A separately encoded content-induced structural view supports this routing space without overwriting observed collaborative topology.

\rev{Across four Amazon domains and the non-Amazon MicroLens-100K benchmark, cell-wise tests of seed-wise performance differences support MAGNET's gains over the strongest protocol-compatible non-MAGNET baselines in the main comparisons, while MAGNET achieves the highest descriptive means in the reported tail-item and low-history warm-start user evaluation.} In matched expert controls, \rev{both structured variants achieve higher five-seed mean NDCG@20 than homogeneous and free-mixture alternatives on all five datasets; fixed templates provide a strong shared default on the four Amazon domains, while the anchored variant holds a small numerical edge on MicroLens.} On the four Amazon domains, anchor-group routing is also descriptively consistent with independently defined diagnostic evidence regimes. Together, these findings support a structural perspective on multimodal fusion: the appropriate evidence mixture is interaction dependent, while structured coverage provides a stronger empirical default than unconstrained flexibility in the evaluated settings.

\paragraph{Limitations and future directions.}
While our empirical findings consistently support the proposed framework, several boundaries should be noted. \rev{First, the conclusions condition on the shared pre-extracted visual and textual feature package, and the Dom/Bal/Com families constitute a compact coverage basis rather than an exhaustive taxonomy.} Second, because user cues are pooled from the complete training-only history, an observed training positive contributes $1/|\mathcal I_u|$ to its historical profile; evaluating leave-one-out cue construction remains future work. \rev{Third, the reported seed-wise difference tests quantify optimization-seed variability under a fixed interaction split rather than split-to-split uncertainty, and the user-subgroup evaluation reflects a low-history warm-start regime rather than zero-interaction cold start.} Future work will explore end-to-end multimodal fine-tuning, robust routing under missing or corrupted modalities, and broader non-commerce recommendation scenarios.

% \begin{acks}
% This work was supported by Beijing Natural Science Foundation (JQ24019). This work was supported in part by the National Natural Science Foundation of China (No. 62576047, 52572349). This work is being supported by the Open Fund of the Key Laboratory for Civil Aviation Collaborative Air Traffic Management Technology and Applications (No. 2025-001) and sponsored by SMP-Z Large Model Fund (No. CIPS-SMP20250313).
% \end{acks}

% Preserve the ACM bibliography font while avoiding a nearly empty final page.
\setlength{\bibsep}{0pt plus 0.3ex}
\bibliographystyle{ACM-Reference-Format}
\bibliography{ref}

\end{document}